\documentclass[twoside,11pt]{article}
\usepackage{jair, theapa, rawfonts}
\usepackage[utf8]{inputenc}
\usepackage[T1]{fontenc}
\usepackage{graphicx}
\usepackage{subfigure}
\usepackage{amsmath}
\usepackage{amsthm}
\usepackage{amssymb}
\usepackage{IEEEtrantools}
\newtheorem{definition}{Definition}
\usepackage{bbm}
\newtheorem{theorem}{Theorem}
\newtheorem{lemma}{Lemma}
\newtheorem{assumption}{Assumption}
\DeclareMathOperator{\sign}{sign}
\DeclareMathOperator{\argmin}{argmin}
\DeclareMathOperator{\argmax}{argmax}
\DeclareMathOperator{\clustering}{clustering}
\newtheorem{remark}{Remark}

\newtheorem{problem}{Problem}

\usepackage{tikz}
\usetikzlibrary{shapes.symbols}
\usepackage{fontawesome}
\usepackage{algorithm}
\usepackage{algpseudocode}
\usepackage{booktabs}
\usepackage{multirow}
\usepackage{xurl}

\ShortHeadings{Geometrically Inspired Kernel Machines for Collaborative Learning}
{Kumar, Valentinitsch, Fuchs, Brucker, Bowles, Husakovic, Abbas, and Moser}
\firstpageno{1}

\begin{document}

\title{Geometrically Inspired Kernel Machines for Collaborative Learning Beyond Gradient Descent}

\author{\name Mohit Kumar \email mohit.kumar@uni-rostock.de \\
       \addr Faculty of Computer Science and Electrical Engineering \\
       University of Rostock, Germany\\
        and\\
       \addr Software Competence Center Hagenberg GmbH       \\   
       A-4232 Hagenberg, Austria
       \AND
       \name Alexander Valentinitsch \email alexander.valentinitsch@scch.at \\
       \name Magdalena Fuchs \email magdalena.fuchs@scch.at \\
       \name Mathias Brucker \email mathias.brucker@scch.at \\
               \addr Software Competence Center Hagenberg GmbH\\
                 A-4232 Hagenberg, Austria
                 \AND
         \name Juliana Bowles \email jkfb@st-andrews.ac.uk  \\
         \addr School of Computer Science \\
         University of St Andrews, UK \\
         and\\
           \addr Software Competence Center Hagenberg GmbH\\
                 A-4232 Hagenberg, Austria \\
            \name Adnan Husakovic \email adnan.husakovic@primetals.com  \\
         \name Ali Abbas \email ali.abbas@primetals.com  \\
         \addr Primetals Technologies Austria GmbH \\
                 A-4031 Linz, Austria   
                  \AND
       \name Bernhard A. Moser \email bernhard.moser@scch.at \\
         \addr Institute of Signal Processing \\
         Johannes Kepler University Linz, Austria \\
                 and\\
                \addr Software Competence Center Hagenberg GmbH\\
                 A-4232 Hagenberg, Austria
}


\maketitle

\begin{abstract}
This paper develops a novel mathematical framework for collaborative learning by means of \emph{geometrically inspired kernel machines} which includes statements on the bounds of generalisation and approximation errors, and sample complexity. For classification problems, this approach allows us to learn bounded geometric structures around given data points and hence solve the global model learning problem in an efficient way by exploiting convexity properties of the related optimisation problem in a Reproducing Kernel Hilbert Space (RKHS). In this way, we can reduce classification problems to determining the closest \emph{bounded geometric structure} from a given data point. Further advantages that come with our solution is that our approach does not require clients to perform multiple epochs of local optimisation using stochastic gradient descent, nor require rounds of communication between client/server for optimising the global model. We highlight that numerous experiments have shown that the proposed method is a competitive alternative to the state-of-the-art.
\end{abstract}

\section{Introduction}
\label{Introduction}
In different domains of modern industry such as iron- or steelmaking, substantial amounts of data is generated and fused together to describe complex processes. These data are often distributed from multiple data vendors and clients, making collaborative data analysis and model training challenging. A potential bottleneck of using traditional centralized machine learning approaches is the overall data aggregation to a single location. The centralization of the data is not always feasible due to privacy concerns and logistical constraints.

Our focus here is on collaborative learning from distributed and privately owned data. Federated learning is an increasingly popular approach to collaborative learning between multiple clients without the need to exchange raw training data. Given this advantage, federated learning can play a crucial role in process industry by leveraging distributed data to improve model performance while preserving data privacy. The classical federated learning approach~\shortcite{pmlr-v54-mcmahan17a,DBLP:conf/mlsys/LiSZSTS20,10.5555/3495724.3496362} aims to train a common global model by repeating the following two operations: 1) training client local models using local data, and 2) aggregating these local models to update a global model. However, the sampling of distributed data from different local distributions makes it challenging to design and analyse efficient federated learning algorithms in practice~\shortcite{ye2023heterogeneous,morafah2022rethinking}. In general, essential requirements for federated learning include: 1) the capability of addressing heterogeneity among local data distributions, 2) the support for communication efficiency (allowing clients to transfer the required amount of parameters to the server under limited communication bandwidth), and 3) overall computational efficiency (for real-time operations)~\shortcite{kairouz_advances_2021}.
\subsection{Central Problem}
In order to develop an accurate collaborative learning method that is efficient in both communication and computation, we formulate the following research questions:
\begin{itemize}
\item[] {\em Q1: Can we build a theoretical analysis framework for collaborative learning from distributed and statistically heterogeneous data that, without making any assumptions on data distributions, allows us to calculate in practice: 1) the generalisation and approximation error bounds, and 2) the minimum number of training samples required to reduce the risk in approximating the target function below $\epsilon$ (for any $\epsilon > 0$) with probability at least $1-\delta$ (for any $\delta \in (0,1)$)?}
\item[] {\em Q2: Can we solve the global model optimisation problem in the federated setting without requiring multiple rounds of communication between clients and the server?}
\item[] {\em Q3: Can kernel machine learning theory provide a competitive and computationally efficient alternative to stochastic gradient descent-based optimisation in a federated setting?}
\end{itemize}
\subsection{The State of the Art}
\begin{figure*}[!h]
\centerline{\subfigure[Local dataset 1: A set of samples has been labelled into 3 classes and each class modelled through a KAHM. The image of each KAHM defines a bounded geometric structure around data points.]{\includegraphics[width=0.3\textwidth]{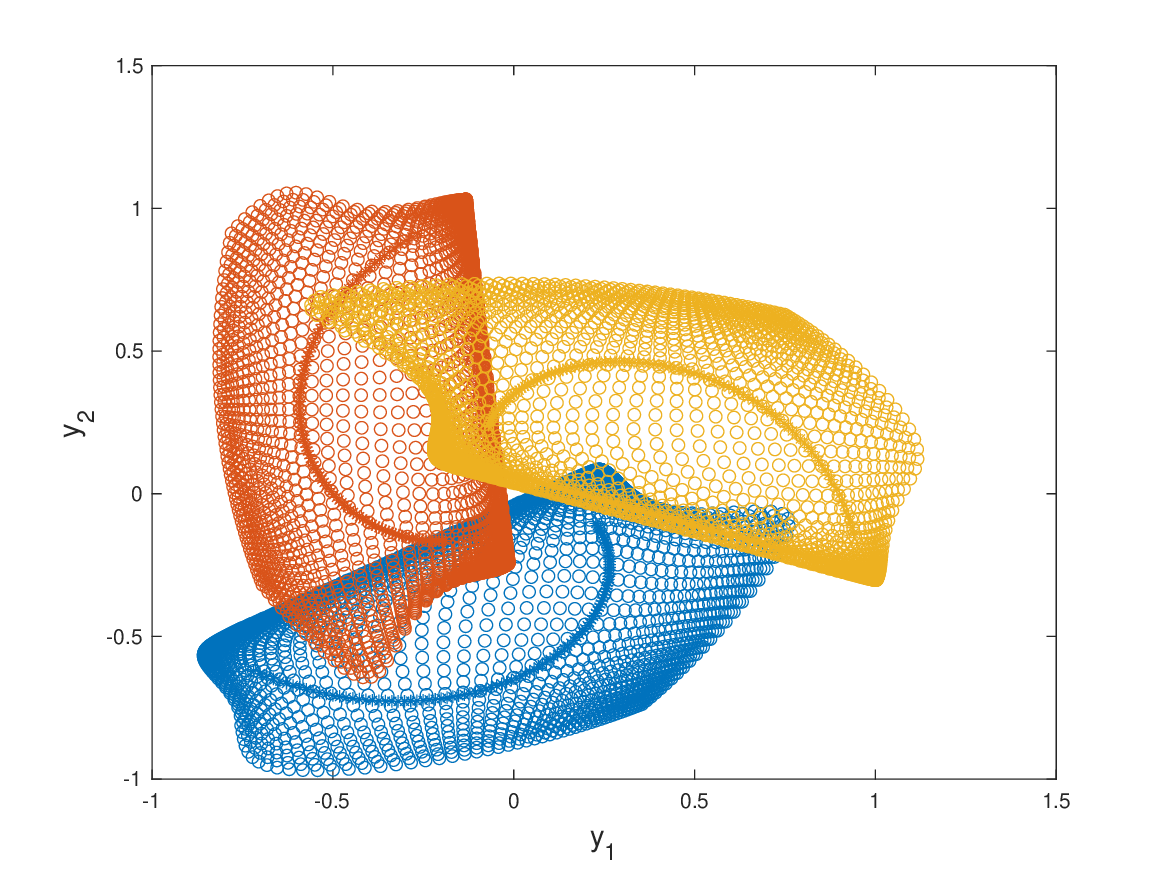}\label{fig_KAHMs:1}} \hfil 
\subfigure[Local dataset 2: Another set of samples also labelled into 3 classes with classes modelled through a KAHM. As before, the image of each KAHM defines a bounded geometric structure around data points.]{\includegraphics[width=0.3\textwidth]{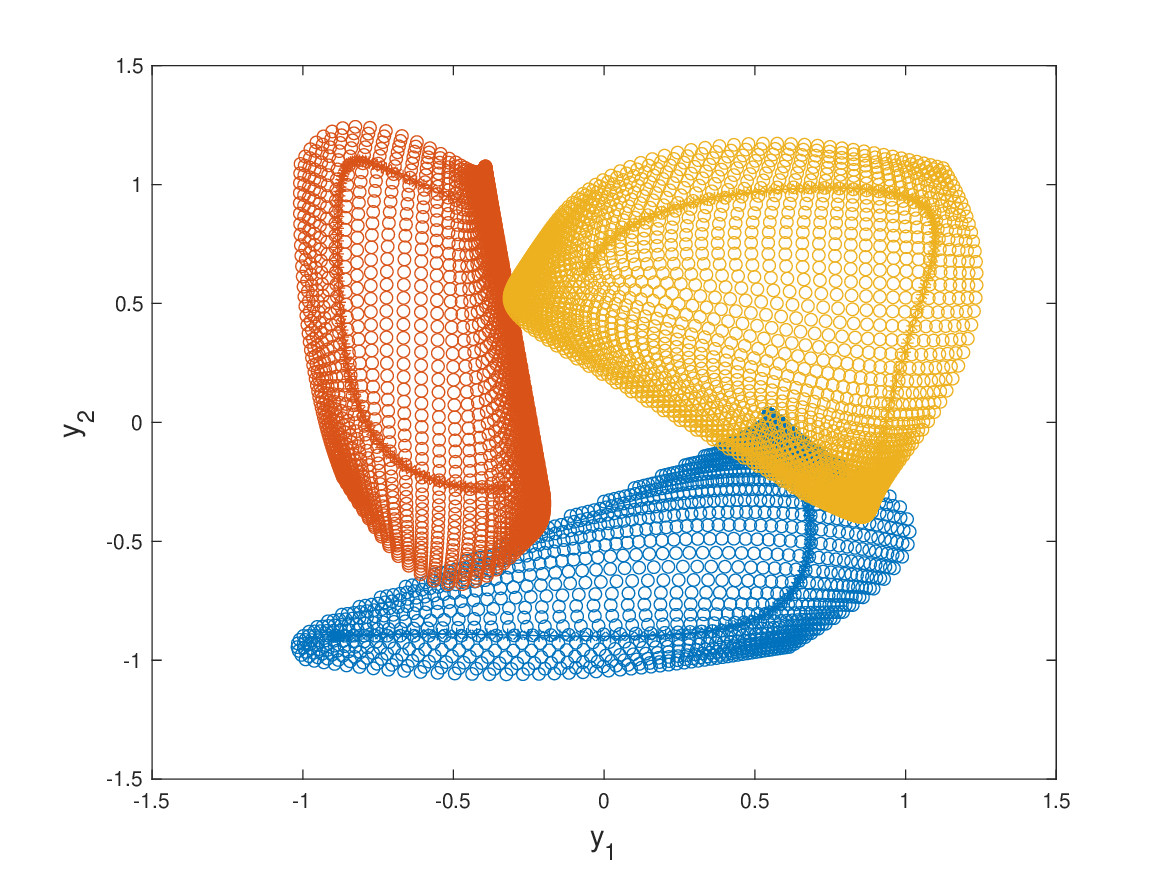}\label{fig_KAHMs:2}} \hfil 
\subfigure[Global KAHM: The local class-specific KAHMs, trained independently using dataset 1 and 2, can be aggregated to build the global class-specific KAHM using the distance measure.]{\includegraphics[width=0.3\textwidth]{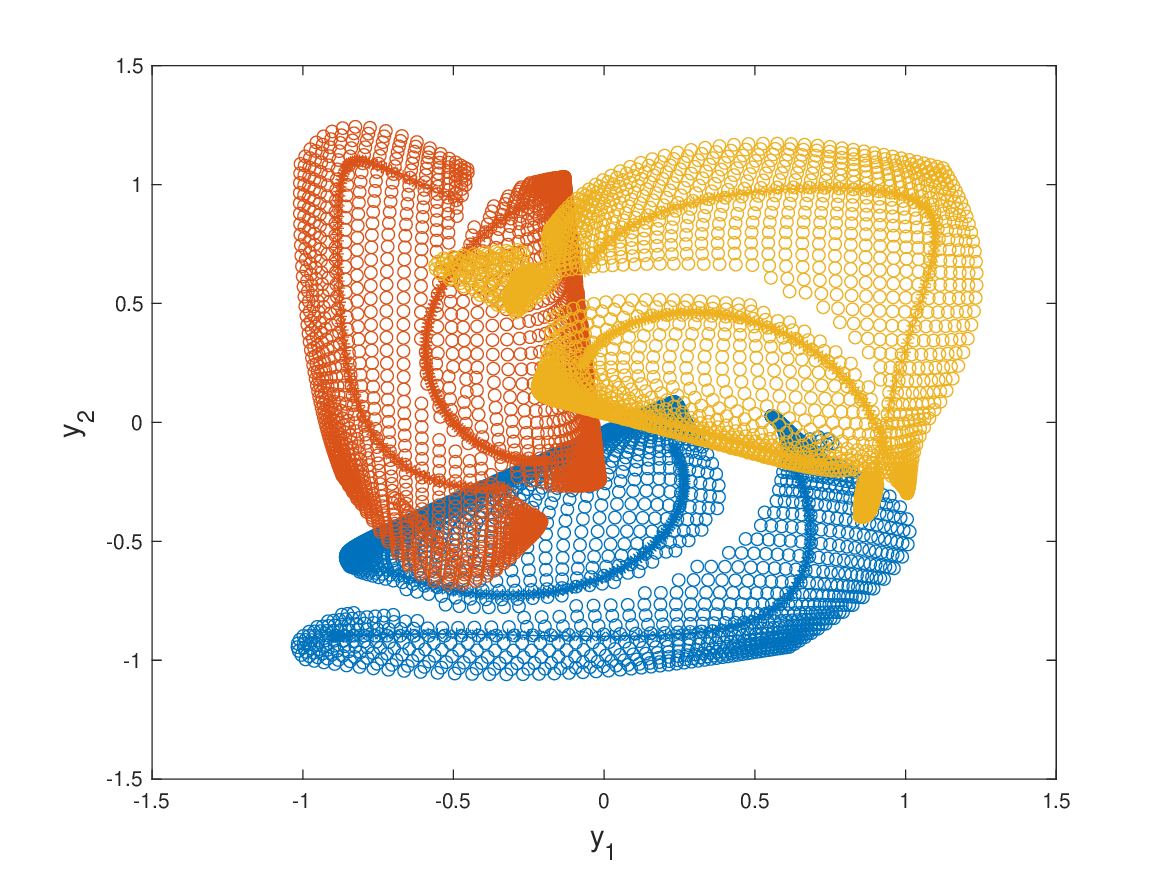}\label{fig_KAHMs:3}}}
\centerline{\subfigure[Decision boundary of classifier 1: the KAHM induced distance measure allows us to build a classifier from local dataset 1.]{\includegraphics[width=0.3\textwidth]{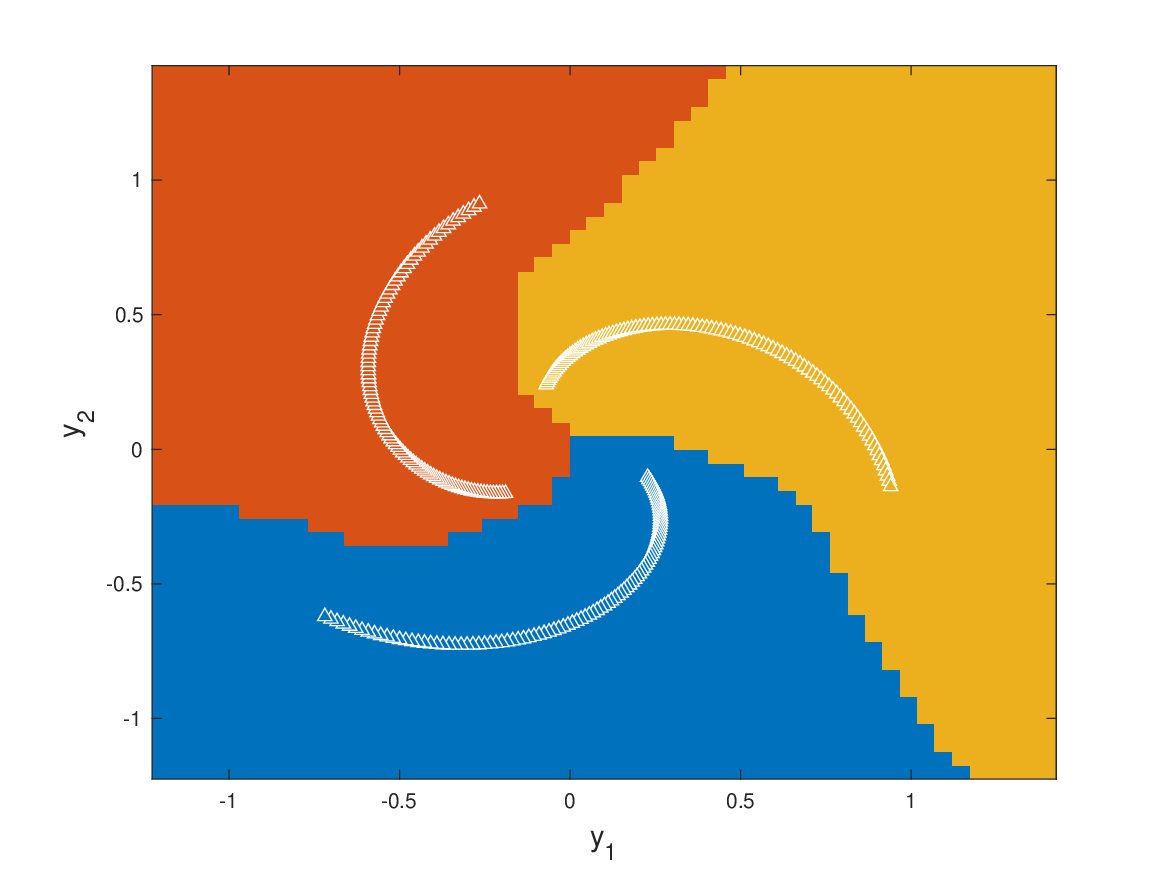}\label{fig_KAHMs:4}} \hfil 
\subfigure[Decision boundary of classifier 2: the KAHM induced distance measure also allows us to build a classifier from local dataset 2.]{\includegraphics[width=0.3\textwidth]{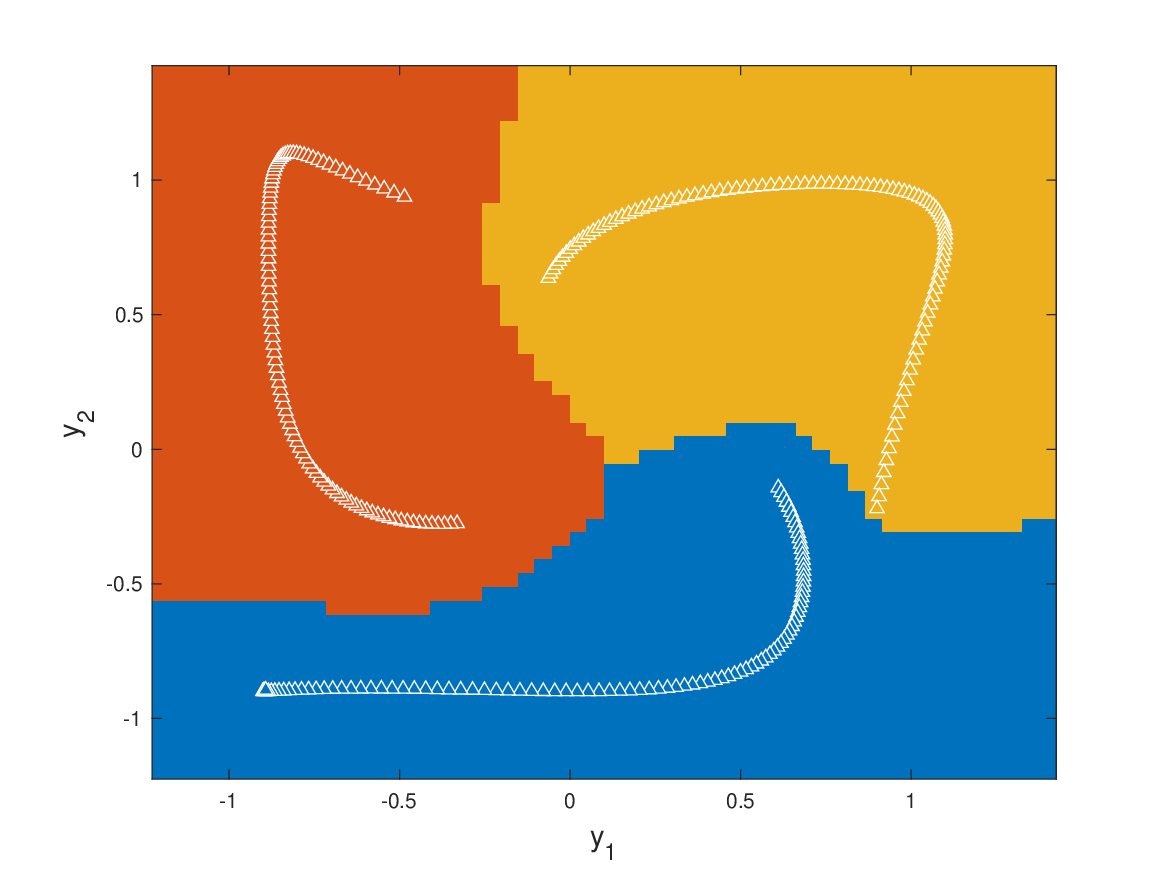}\label{fig_KAHMs:5}} \hfil 
\subfigure[Decision boundary of a global classifier: The global KAHMs make it possible for us to obtain a global classifier from the distributed datasets 1 and 2 in a federated setting.]{\includegraphics[width=0.3\textwidth]{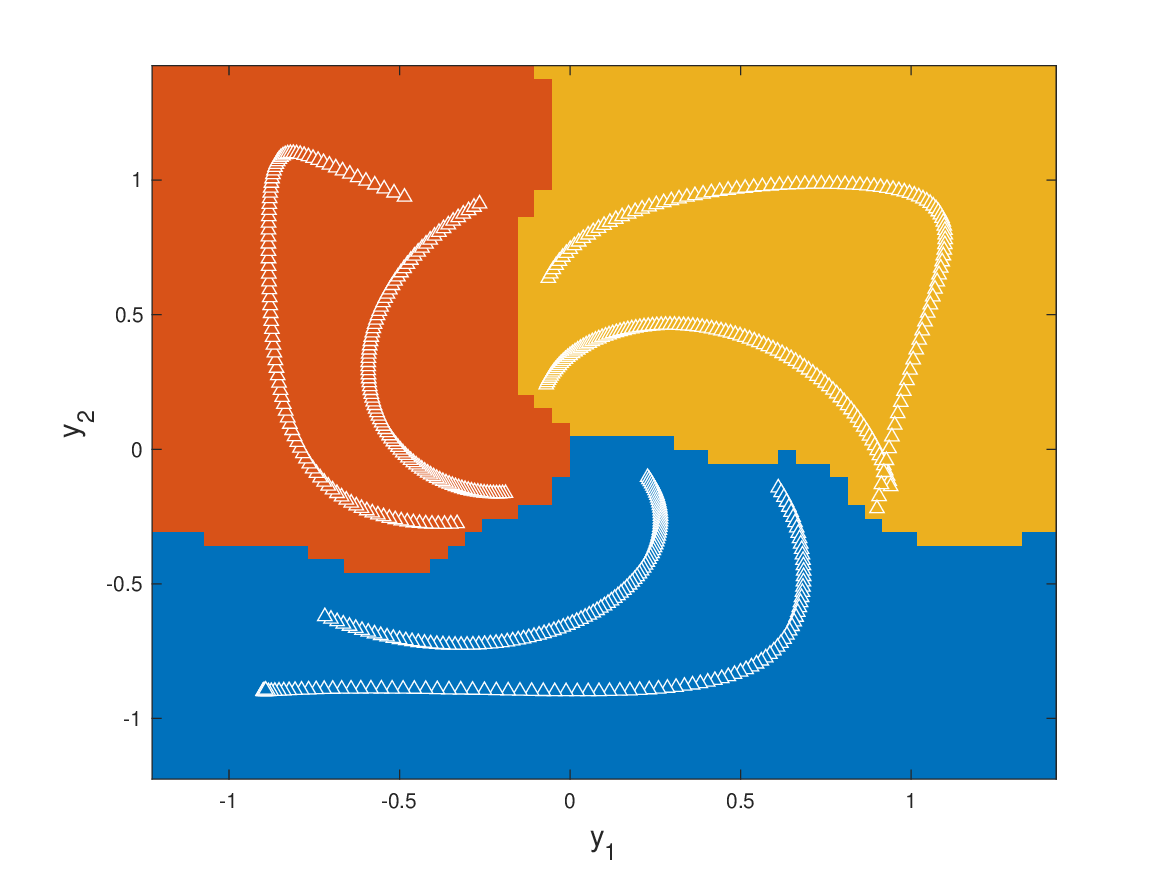}\label{fig_KAHM:6}}}
\caption{An illustration of the KAHM-based learning from distributed data.}
\label{fig_KAHMs}
\end{figure*}

\paragraph{Addressing Data Heterogeneity.} The issue of data heterogeneity in federated learning has been previously addressed by learning a personalised model for each client assuming that data features share a common global representation, while statistical heterogeneity across clients is attributed to the labels \shortcite{pmlr-v139-collins21a}. The personalised federated learning problem has been also studied under the model-agnostic meta-learning framework with the goal of finding an initial shared model that can easily be adapted to local datasets by performing a few steps of gradient descent~\cite{NEURIPS2020_24389bfe}. Another personalised approach is that clients, instead of fully utilising the averaged global parameters for initialisation, only select a subset of the global model’s parameters, and load the remaining parameters from previous local models \shortcite{NEURIPS2021_c429429b}. Adversarial learning is another approach to deal with heterogeneous data features, where a discriminator is trained to distinguish the representations of the clients, while the clients aim to generate indistinguishable representations \cite{pmlr-v202-li23j}. Alternatively, a clustered federated learning approach has been proposed based on the grouping of clients into clusters so that clients of the same cluster share the same model \shortcite{9174890,10.1609/aaai.v37i8.26197}. 
\paragraph{Theoretical Analysis Framework.} In federated learning, the underlying models can be chosen from a reproducing kernel Hilbert space \cite{9625795,ghari2022personalized} allowing for an application of the powerful kernel theory for  design and analysis. Kernels have been applied in machine learning over the years~\shortcite{10.1214/009053607000000677,NIPS2017_05546b0e} and have recently gained renewed attention. In particular, the parallels between the properties of deep neural networks and kernel methods have been established to indicate that some key phenomena of deep learning are manifested similarly in kernel methods in the {\em overfitted} regime~\cite{pmlr-v80-belkin18a}, and deep kernel machines have been introduced~\shortcite{pmlr-v51-wilson16,10.1007/s10044-020-00933-1}. Kernel-based models are effective for learning representations~\shortcite{7477690,KAMPFFMEYER2018816,pmlr-v89-laforgue19a} and facilitate analytical solutions for learning problems using a broad range of mathematical techniques. A convergence guarantee for federated learning can be established for strongly convex and smooth objective functions \shortcite{Li2020On,pmlr-v108-bayoumi20a,DBLP:journals/jair/QuLLZZ23}. For one-hidden layer neural network with ReLU activations, an analysis of federated learning can be provided \shortcite{li2021fedbn} by describing the training dynamics of federated learning by means of Neural Tangent Kernel (NTK)~\shortcite{10.5555/3327757.3327948}. The gradient descent training dynamics of artificial neural networks follows that of the gradient descent of the functional cost with respect to a kernel: NTK~\shortcite{10.5555/3327757.3327948}. A NTK based framework makes use of the theory on over-parameterised neural networks to provide proof of convergence of gradient descent and generalisation bound for over-parameterized ReLU neural networks in federated learning~\shortcite{pmlr-v139-huang21c}. 
\paragraph{Variational Optimisation as an Alternative to Gradient Descent.} A kernel-based approach that does not rely on gradient descent-based learning and instead uses variational optimisation for deriving analytically the learning solutions, has been previously studied \shortcite{8888203,9216097,ZHANG2022128,KUMAR20211,ZHANG2023120145,10.1007/978-3-030-87101-7_13,10.1007/978-3-030-87101-7_14,10012502}. This kernel-based variational optimisation approach was considered for privacy-preserving learning under the differential privacy framework \shortcite{kumar2023differentially,KUMAR202187,10.1145/3386392.3399562} and fully homomorphic encryption \shortcite{10012502}, and can potentially be explored for federated learning as well. So far there have been no attempts to extend the variational optimisation approach to federated learning in such a way that it can handle all our research questions Q1, Q2, and Q3. 
\paragraph{Geometrically Inspired Kernel Approach as an Efficient Alternative.} For collaborative learning in a federated setting, a geometrically inspired kernel approach has been introduced~\cite{KAHM,kumar2023secure}. A recent paper~\cite{KAHM} introduced a so-called {\em Kernel Affine Hull Machine (KAHM)}, wherein a representation of given data points is learned in RKHS to define a bounded geometric structure around data points within the affine hull of data points. The KAHM makes it possible to compute at any arbitrary point a measure of its distance from the data samples. The significance of this is that the KAHM's induced distance measure cannot only be used for classification, but for federated learning by aggregating locally trained KAHMs to build a global KAHM. Note that the crucial significance of KAHMs for learning from distributed data comes from the fact that a {\em global model can be built by aggregating local models simply using a distance measure without requiring gradient-based learning of the global model parameters}. This is best illustrated through an example as shown in Fig.~\ref{fig_KAHMs}. Consequently, a KAHM-based approach is computationally more efficient and indeed promising for federated learning~\cite{kumar2023secure,KAHM}. Moreover, KAHMs can be used to mitigate the accuracy-loss issue of differential privacy, where the post-processing property of differential privacy is leveraged for fabricating new data samples by means of a geometric model ensuring that the geometric modelling error of fabricated data samples is never larger than that of original data samples while simultaneously achieving the privacy-loss bound. Although a mathematical proof of fabricated data samples with modelling error less than that of original data samples is provided \cite{KAHM}, no generalisation error analysis and performance guarantees have been provided for the KAHM-based learning method. The federated learning solution, as suggested in \cite{KAHM,kumar2023secure}, has been introduced in a rather {\em adhoc} manner without providing a mathematical theory to justify the solution.
\begin{remark}[Research Gap]\label{remark_researck_gap}
Existing studies on federated learning address data heterogeneity 
through a personalised or clustered approach, develop a theoretical analysis framework for gradient descent-based learning, and introduce a KAHM-based efficient solution without providing a mathematical theory as a suitable justification. However, these aspects have been studied separately and not together in a unified manner. Indeed, there is a lack of a unified collaborative learning framework that is powerful enough to simultaneously address our formulated research questions Q1, Q2, and Q3.  
\end{remark}
\subsection{Contributions}
In this study, we give an affirmative answer to Q1, Q2, and Q3 by providing a novel approach (based on KAHMs) that harnesses the distributed computational power across clients. We provide a unified framework for the design and analysis of collaborative learning by means of geometrically inspired kernel machines such as KAHMs. Our contributions are summarised in the following four aspects:
\paragraph{Theoretical Framework:} We develop a framework to analyse KAHM-based collaborative learning in a federated setting. We introduce a {\em novel kernel function defined by a global KAHM (that aggregates local KAHMs) such that the kernel function evaluates the degree of similarity between two data points in terms of their distance from training data samples}. The hypothesis space for the learning is suitably (specifically, a convex hull) defined in the RKHS associated to the novel kernel function. An {\em upper bound on the Rademacher complexity of the hypothesis space} is provided (in Theorem~\ref{result_rademacher_complexity_hypthesis_space}) to derive a {\em uniform bound on the generalisation error} (in Theorem~\ref{result_generalisation_error}).  
\paragraph{Beyond Gradient Descent Learning Regime:} Unlike most studies, we move beyond the gradient descent learning regime to derive a collaborative learning solution, that utilises the idea of KAHM's induced distance measure based aggregation of local geometrical models to build a global geometrical model. Our contribution lies in considering the global model learning problem (in Problem~\ref{problem_learning}) and showing that under a realistic assumption, it is possible to {\em derive analytically a learning solution} (in Theorem~\ref{theorem_learning_solution}) that {\em does not require estimating global model parameters}. The underlying assumption is that there is only a small error in the fitting of training data points by the KAHM. This assumption is realistic and is validated through various experiments as well. {\em Since the learning solution does not require estimating the global model parameters, no rounds of communication between clients and server are required for optimising the global model, and the clients are also not required to perform multiple epochs of local optimisation using stochastic gradient descent}. The advantage of our approach is hence that it leads to a {\em communication and computationally efficient collaborative learning solution}. 
\paragraph{Performance Guarantees:} The generalisation error bound allows us to derive an {\em upper bound on the error in approximating the target function} (in Theorem~\ref{result_generalisation_error_3}). Note that the {\em target function approximation risk bound can be calculated in practice and decays as $\mathcal{O}(1/\sqrt{N})$, where $N$ is the total number of training data samples distributed across multiple clients}. Remarkably, the risk bound depends only on $N$, and thus the {\em sample complexity} (i.e. the number of training samples needed for an arbitrarily small risk in approximating the target function) is calculated (in Lemma~\ref{lemma_result_generalisation_error_4}) and plotted (in Fig.~\ref{fig_sample_complexity}). We additionally provide a deterministic analysis (in Theorem~\ref{theorem_deterministic_analysis}) that further justifies the proposed solution via an interpretation in-terms of distance from training data points.
\paragraph{Competitive Alternative:} A KAHM-based learning approach provides a {\em competitive alternative to the state of the art federated learning methods}, and in fact outperforms traditional solutions. Furthermore, the KAHM approach facilitates and enhances {\em cross-domain knowledge transfer in federated settings}. Experiments are used to show the improved performance of the proposed method when compared to the state of the art methods.              
\subsection{Proposed Approach, Novelty, and Significance}
Our unified approach for the development of a collaborative learning framework for addressing our research questions (Q1, Q2, and Q3) consists of the following 9 steps:    
\paragraph{Step 1: Define a KAHM induced kernel function.} KAHMs let us define a novel kernel function that measures the similarity between two data points in terms of their distance from training samples of a class.
\paragraph{Step 2: Define a data-dependent hypothesis space for learning.} To predict the association between a class-label and a data point, the considered hypothesis space is defined by the given data samples and is in the form of a convex hull within the RKHS associated to the KAHM induced kernel function.   
\paragraph{Step 3: Calculate the upper bound of the Rademacher complexity of the hypothesis space.} The Rademacher complexity of the considered hypothesis space has an upper bound such that this bound can be calculated in practice.
\paragraph{Step 4: Derive the generalisation error bound for the hypothesis space.} Following the standard approach, the Rademacher complexity can be used to derive a uniform bound on the generalisation error.
\paragraph{Step 5: Formulate the global model learning problem.} The global model learning problem can be formulated as an optimisation problem over a {\em suitably} chosen subset of the hypothesis space.
\paragraph{Step 6: Exploit the convex hull form of the hypothesis space for deriving a learning solution analytically.} The convex hull form of the hypothesis space can be leveraged together with a realistic assumption to derive a learning solution that does not require estimating any of the model parameters using gradient descent or any other numerical algorithm.  
\paragraph{Step 7: Derive the upper bound of the error in approximating the target function.} The generalization error bound can be used to derive an upper bound of the error in approximating the target function. 
\paragraph{Step 8: Calculate the sample complexity.} The target function approximation risk bound can be used to calculate the sample complexity.
\paragraph{Step 9. Provide a deterministic analysis of the solution.} The upper bound of the KAHM induced distance function can be used to analyse and interpret the solution in terms of the distance from training data points. 
\begin{remark}[Novelty]\label{remark_290520241955}
The above 9 steps offer a novel approach to the development of a collaborative learning framework. In particular, the introduction of a KAHM-induced kernel function (step 1) and exploiting the convex hull form of the hypothesis space (step 2) for deriving the learning solution analytically (step 6) are original. This is the first study applying geometrically inspired kernel machines (i.e., KAHMs) for a rigorous design and analysis of collaborative learning solutions. 
\end{remark}
\begin{remark}[Significance]\label{remark_30052024064}
The proposed approach has been carefully designed to address the formulated research questions. Q1 is addressed by steps 4, 7, and 8. Q2 and Q3 are addressed by step 6. A new deterministic way of studying and solving the learning problem is provided by step 9. The work provides a new theoretical analysis framework for learning beyond gradient descent.    
\end{remark}
\subsection{Structure of the Paper}
Section~\ref{sec_background} presents the necessary mathematical background underlying our work. Section~\ref{sec_theory} develops the theory for KAHMs and includes steps 1-4 of the proposed approach outlined earlier. Section~\ref{sec_210420241350} continues with steps 5-9, thereby solving the collaborative learning problem. The experimental evaluation of our approach is given in Section~\ref{sec_210420241503}, followed by concluding remarks in Section~\ref{sec_conclusion}.
\section{Mathematical Prerequisites and Notations}\label{sec_background}
This section introduces the notation used throughout, presents the distributed data setting, and provides a review of the notion of KAHM.  
\subsection{Notation}\label{sec}
In this paper, all matrices are denoted using boldface font. The following notation is used:
\begin{itemize}
\item Let $n,p,c,N,Q,C \in \mathbb{Z}_{+}$ be the positive integers. 
\item For a set $\{y^1,\cdots,y^N \} \subset  \mathbb{R}^p$, its affine hull is denoted as $\mathrm{aff}(\{y^1,\cdots,y^N \})$. 
\item For a scalar $a\in \mathbb{R}$, $|a|$ denotes its absolute value. For a set $A$, $|A|$ denotes its cardinality. For a real matrix $\mathbf{Y}$, $\mathbf{Y}^T$ is the transpose of $\mathbf{Y}$. 
\item For a vector $y \in \mathbb{R}^p$, $\| y\|$ denotes the Euclidean norm and $y_j$ (and also $(y)_j$) denotes the $j^{th}$ element. For a matrix $\mathbf{Y}\in \mathbb{R}^{N \times p}$, $\|\mathbf{Y}\|_2$ denotes the spectral norm, $\| \mathbf{Y} \|_F$ denotes the Frobenius norm, $(\mathbf{Y})_{i,:}$ denotes the $i^{th}$ row, $(\mathbf{Y})_{:,j}$ denotes the $j^{th}$ column, and $(\mathbf{Y})_{i,j}$ denotes the $(i,j)^{th}$ element. 
\item Let $\mathcal{X} \subset \mathbb{R}^n$ be a region. A RKHS, $\mathcal{H}_{k}(\mathcal{X})$, is a Hilbert space of functions $f: \mathcal{X} \rightarrow \mathbb{R}$ on a non-empty set $\mathcal{X}$ with a reproducing kernel $k: \mathcal{X} \times \mathcal{X} \rightarrow \mathbb{R}$ satisfying $\forall x \in \mathcal{X}$ and $\forall f \in \mathcal{H}$,
      \begin{IEEEeqnarray}{rCl}
k(\cdot,x) & \in &  \mathcal{H}_k(\mathcal{X}), \\
\langle  f, k(\cdot,x)  \rangle_{\mathcal{H}_k(\mathcal{X})} & = & f(x),
      \end{IEEEeqnarray}
where $\langle \cdot, \cdot \rangle_{\mathcal{H}_k(\mathcal{X})} : \mathcal{H}_k(\mathcal{X}) \times \mathcal{H}_k(\mathcal{X}) \rightarrow \mathbb{R}$ is an inner product on $\mathcal{H}_k(\mathcal{X})$. 
Let $\left \| f \right \|_{\mathcal{H}_k(\mathcal{X})} := \sqrt{\langle f, f \rangle_{\mathcal{H}_k(\mathcal{X})}}$ denote the norm induced by the inner product on $\mathcal{H}_k(\mathcal{X})$. 
\item Let $\mathbf{K}$ be a symmetric matrix, $ \mathbf{K} \succ 0$ denotes that  $\mathbf{K}$ is positive-definite.
\item Let$(\Omega_{y,z,q},\mathcal{F}_{y,z,q},\mu_{y,z,q})$ be a {\em probability space} and $(y,z,q): \Omega_{y,z,q} \rightarrow \mathbb{R}^p \times \{0,1\}^C \times \{1,2,\cdots,Q \}$ be a random vector on $\Omega_{y,z,q}$. Let $\mathcal{B}(\mathbb{R}^p \times \{0,1\}^C \times \{1,2,\cdots,Q \})$ denote the Borel $\sigma-$algebra on $\mathbb{R}^p \times \{0,1\}^C \times \{1,2,\cdots,Q \}$. Let $\mathbb{P}_{y,z,q}: \mathcal{B}(\mathbb{R}^p \times \{0,1\}^C \times \{1,2,\cdots,Q \}) \rightarrow \mathbb{R}$ be the distribution of $(y,z,q)$ given as         
      \begin{IEEEeqnarray}{rCl}
\mathbb{P}_{y,z,q} & := &  \mu_{y,z,q} \circ (y,z,q)^{-1}.
      \end{IEEEeqnarray}    
\end{itemize}
\subsection{Distributed Data Setting}
We consider the multi-class problem, where a label vector $z \in \{0,1\}^C$ with $C \geq 2$ is assigned to a data point $y \in \mathbb{R}^p $ such that the $c-$th element of vector $z$, $z_c \in \{0,1\}$, represents the association of $y$ with the $c-$th class. We consider the problem of collaborative learning from distributed data, where a number of clients $Q$ $(Q \geq 1)$  participate in the learning. Let $q \in \{1,2,\cdots,Q \}$ be the client characterising variable. Let $\mathcal{D}$ be a set consisting of $N$ number of samples drawn i.i.d. according to the distribution $\mathbb{P}_{y,z,q} $:  
      \begin{IEEEeqnarray}{rCCCl}
\label{eq_151220231703}   \mathcal{D} & := &  \left\{(y^i,z^i,q^i) \in \mathbb{R}^p \times \{ 0,1\}^C \times \{1,2,\cdots,Q \}  \; \mid \;i \in \{1,2,\cdots,N\}  \right \} & \sim &(\mathbb{P}_{y,z,q})^N.
      \end{IEEEeqnarray} 
Let $\mathcal{I}^{c,q}$ be the set of indices of those samples in the sequence $\left(\left(y^i,z^i,q^i\right) \in \mathcal{D} \right)_{i=1}^N$ which are $c^{th}$ class labelled and owned by client $q$, i.e.,
    \begin{IEEEeqnarray}{rCl}
 \label{eq_040120241522}    \mathcal{I}^{c,q} & : = & \left \{ i \in \{1,2,\cdots,N\} \; \mid \; (z^i)_c = 1,\; q^i = q  \right \}.
      \end{IEEEeqnarray} 
Let $(\mathrm{I}_1^{c,q},\cdots,\mathrm{I}_{|\mathcal{I}^{c,q}|}^{c,q})$ be the sequence of elements of $\mathcal{I}^{c,q}$ in ascending order, i.e., 
\begin{IEEEeqnarray}{rCl}
\mathrm{I}_1^{c,q} & = & \min(\mathcal{I}^{c,q}), \\
\mathrm{I}_i^{c,q} & = & \min(\mathcal{I}^{c,q} \setminus \{\mathrm{I}_1^{c,q}, \cdots, \mathrm{I}_{i-1}^{c,q}\} ),
\end{IEEEeqnarray}
for $i \in \{2,\cdots, |\mathcal{I}^{c,q}|\}$. Let $\mathbf{Y}^{c,q} \in \mathbb{R}^{ |\mathcal{I}^{c,q}| \times p}$ be the matrix storing the $c^{th}$ class labelled and $q^{th}$ client owned samples, i.e., 
\begin{IEEEeqnarray}{rCl}
\label{eq_180420241115}\mathbf{Y}^{c,q} &=& \left[\begin{IEEEeqnarraybox*}[][c]{,c/c/c,} y^{\mathrm{I}_1^{c,q}} & \cdots & y^{\mathrm{I}_{|\mathcal{I}^{c,q}|}^{c,q}} \end{IEEEeqnarraybox*} \right]^T.
\end{IEEEeqnarray}
\begin{remark}[Addressing Statistical Heterogeneity]\label{rem_distributed_data}
Our analysis takes into account the heterogeneity among client's data distributions by automatically considering, for arbitrary clients $q^i$ and $q^j$ with $i \neq j$, the following cases as well:
 \begin{IEEEeqnarray}{rCl}
 \mathbb{P}_{y,z| q}(\cdot,\cdot | q = q^i) & \neq &  \mathbb{P}_{y,z| q}(\cdot,\cdot | q = q^j) ,\\
 \mathbb{P}_{z| y, q}(\cdot | y, q = q^i) & \neq &  \mathbb{P}_{z| y, q}(\cdot | y, q = q^j).
 \end{IEEEeqnarray}
Hence, data shifts amongst clients are being considered.
\end{remark}

\subsection{A Review of Kernel Affine Hull Machines}
We recall the notion of KAHM as originally defined
in~\cite{KAHM}.
\begin{definition}[Kernel Affine Hull Machine (KAHM)~\cite{KAHM}]\label{def_affine_hull_model}
Given a finite number of samples: $\mathbf{Y} = \left[\begin{IEEEeqnarraybox*}[][c]{,c/c/c,} y^1 & \cdots & y^N \end{IEEEeqnarraybox*} \right]^T$ with $y^1,\cdots,y^N \in \mathbb{R}^p$ and a subspace dimension $n \leq p$; a kernel affine hull machine $\mathcal{A}_{\mathbf{Y},n}: \mathbb{R}^p \rightarrow \mathrm{aff}(\{y^1,\cdots,y^N \})$ maps an arbitrary point $y \in \mathbb{R}^p$ onto the affine hull of $\{y^1,\cdots,y^N\}$. 
\end{definition}
The complete definition of $\mathcal{A}_{\mathbf{Y},n}$ is provided in Appendix A. In addition, the subspace dimension $n \leq p$ is practically determined for the given samples using a procedure given in Appendix B. 
\begin{remark}[KAHM for Automated Machine Learning (AutoML)]\label{rem_130520241037}
A KAHM $\mathcal{A}_{\mathbf{Y},n}$, with the choice of subspace dimension $n$ as suggested in Appendix B, is completely defined by the data samples $Y$ without involving any free parameters to be tuned, allowing us to define  an AutoML approach by means of KAHMs. Consequently, in what follows we use $\mathcal{A}_{\mathbf{Y}}$ to denote a KAHM.
\end{remark}
\begin{theorem}[KAHM as a Bounded Function~\cite{KAHM}]\label{result_kahm_bounded_function}
The KAHM $\mathcal{A}_{\mathbf{Y}}$,  associated to $\mathbf{Y} = \left[\begin{IEEEeqnarraybox*}[][c]{,c/c/c,} y^1 & \cdots & y^N \end{IEEEeqnarraybox*} \right]^T$ with $y^1,\cdots,y^N \in \mathbb{R}^p$,  is a bounded function on $\mathbb{R}^p$ such that for any $y \in \mathbb{R}^p$, 
\begin{IEEEeqnarray}{rCl}
\label{eq_100120231400} \| \mathcal{A}_{\mathbf{Y}}(y)\| & < &   \left\|\mathbf{Y} \right \|_2\left(1 + \frac{pN^2}{2\|\mathbf{Y}\|_F^2} \right).
\end{IEEEeqnarray}  
Thus, the image of $\mathcal{A}_{\mathbf{Y}}$ is bounded such that 
 \begin{IEEEeqnarray}{rCl}
 \mathcal{A}_{\mathbf{Y}}[\mathbb{R}^p]& \subset &\left\{ y \in \mathbb{R}^p \; \mid \;  \| y\| < \left\|\mathbf{Y} \right \|_2 \left(1 + \frac{pN^2}{2\|\mathbf{Y}\|_F^2} \right) \right \}.
   \end{IEEEeqnarray} 
\end{theorem} 
\begin{definition}[A Distance Function Induced by KAHM~\cite{KAHM}]\label{def_distance_function_KAHM}
Given a KAHM $\mathcal{A}_{\mathbf{Y}}$, the distance of an arbitrary point $y\in \mathbb{R}^p$ from its image under $\mathcal{A}_{\mathbf{Y}}$ is given as
 \begin{IEEEeqnarray}{rCl}
 \Gamma_{\mathcal{A}_{\mathbf{Y}}}(y) & := & \left \| y - \mathcal{A}_{\mathbf{Y}}(y) \right \|.
   \end{IEEEeqnarray}    
\end{definition}
\begin{theorem}[$\Gamma_{\mathcal{A}_{\mathbf{Y}}}(\cdot)$ as a Measure of Distance from Data Points~\cite{KAHM}]\label{result_ratio_distances}
The ratio of the distance of a point $y\in \mathbb{R}^p$ from its image under $\mathcal{A}_{\mathbf{Y}}$ to the distance of $y$ from $\{y^1,\cdots,y^N\}$ evaluated as $\left \|\left[\begin{IEEEeqnarraybox*}[][c]{,c/c/c,} y - y^1 & \cdots & y - y^N \end{IEEEeqnarraybox*} \right] \right \|_2$ remains upper bounded as
 \begin{IEEEeqnarray}{rCl}
\label{eq_100120231432} \frac{ \Gamma_{\mathcal{A}_{\mathbf{Y}}}(y) }{\left \|\left[\begin{IEEEeqnarraybox*}[][c]{,c/c/c,} y - y^1 & \cdots & y - y^N \end{IEEEeqnarraybox*} \right] \right \|_2} & < &   1 + \frac{p N^2}{2 \|\mathbf{Y}\|_F^2}.
   \end{IEEEeqnarray}     
   \end{theorem}
Theorem~\ref{result_ratio_distances} states that if a point $y$ is close to points $\{y^1,\cdots,y^N\}$, then the value $\Gamma_{\mathcal{A}_{\mathbf{Y}}}(y)$ cannot be large. Thus, a large value of the distance function at a point $y$ indicates that $y$ must be at a far distance from $\{y^1,\cdots,y^N\}$. 
\subsection{Combining Locally Distributed KAHMs to build a Global KAHM using the Distance Function}
We consider the scenario where 
class labelled data samples are distributed amongst $Q$ different clients. Given $Q$ different KAHMs $\mathcal{A}_{\mathbf{Y}^{c,1}},\cdots,\mathcal{A}_{\mathbf{Y}^{c,q}}$ built independently using data matrices $\mathbf{Y}^{c,1},\cdots,\mathbf{Y}^{c,q}$ respectively, a possible way to combine together the KAHMs is as follows:
 \begin{IEEEeqnarray}{rCl}
 \mathcal{G}_c(y) & = &
 \mathcal{A}_{\mathbf{Y}^{c,\hat{q}(y)}}(y)\\
\hat{q}(y) & = & \mathop{\argmin}_{q \in \{1,2,\cdots,Q \}}  \Gamma_{\mathcal{A}_{\mathbf{Y}^{c,q}}}(y),
   \end{IEEEeqnarray} 
where $\mathcal{G}_c$ is the global KAHM (that combines together the individual KAHMs) and $\Gamma_{\mathcal{A}_{\mathbf{Y}^{c,q}}}$ is the distance function induced by $\mathcal{A}_{\mathbf{Y}^{c,q}}$.
\begin{figure}
\centerline{\subfigure[images of two different KAHMs built independently using two different datasets]{\includegraphics[width=0.4\textwidth]{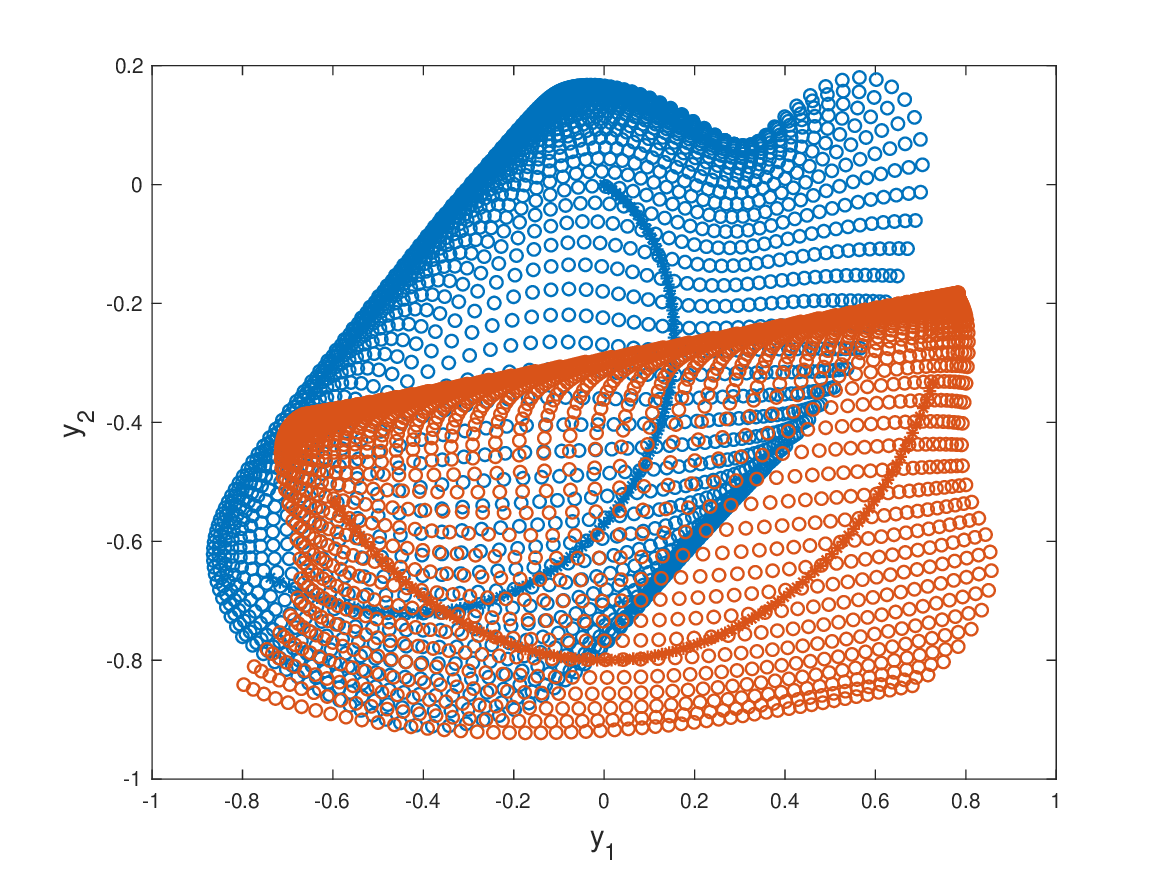}\label{demoGlobalKAHM_1}} \hfil 
\subfigure[image of the global KAHM combining together two independently built KAHMs]{\includegraphics[width=0.4\textwidth]{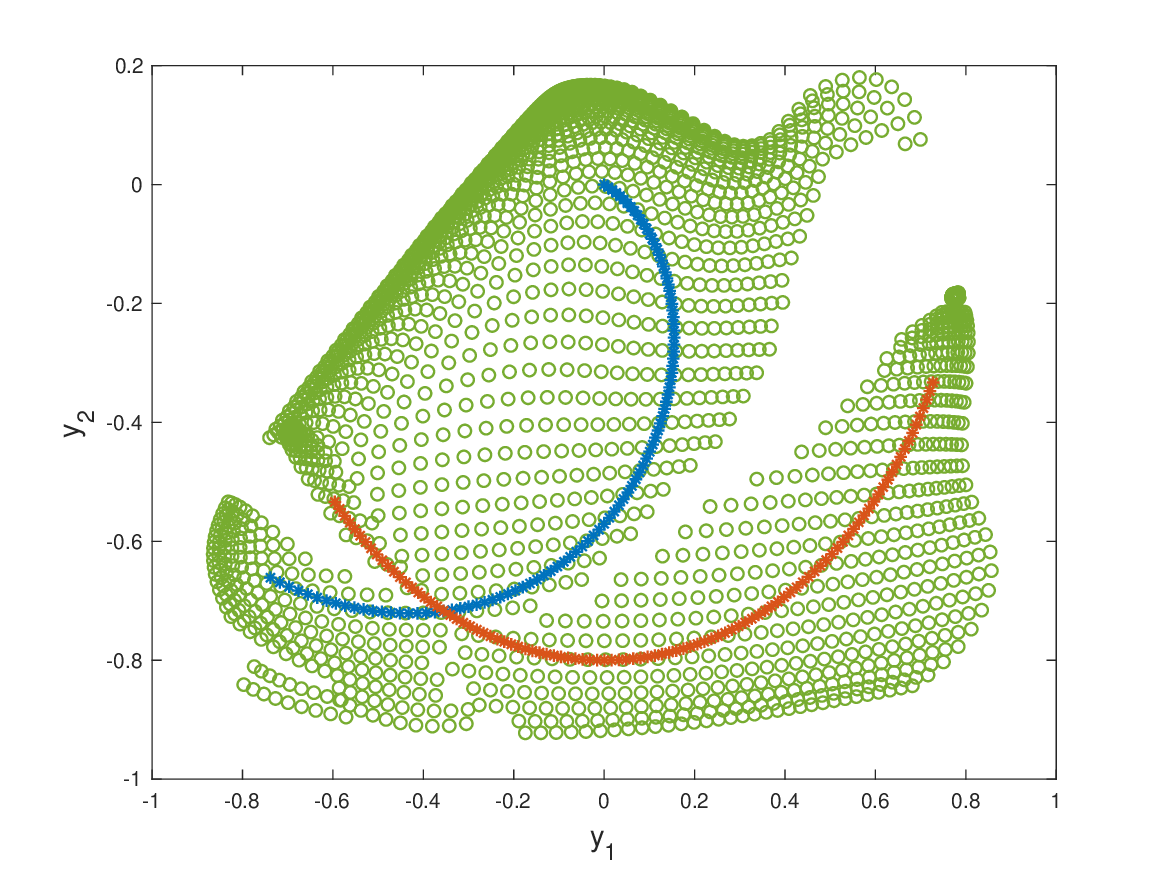}\label{demoGlobalKAHM_2}}}
\caption{An example of combining together local KAHMs to build a global KAHM.}
\label{fig_demo_combination}
\end{figure}
A 2-dimensional data example where two different KAHMs are combined to build a global KAHM is provided in Figure~\ref{fig_demo_combination}. Figure~\ref{fig_demo_combination} shows the images of individual KAHMs (in Figure~\ref{demoGlobalKAHM_1}) and the image of global KAHM (in Figure~\ref{demoGlobalKAHM_2}). 
\begin{definition}[A Distance Function Induced by Global KAHM]\label{def_distance_function_GKAHM}
Given a global KAHM ($\mathcal{G}_c$) that combines together $Q$ local KAHMs ($ \mathcal{A}_{\mathbf{Y}^{c,1}},\cdots, \mathcal{A}_{\mathbf{Y}^{c,q}}$), the distance of an arbitrary point $y\in \mathbb{R}^p$ from its image under $\mathcal{G}_c$ is given as
 \begin{IEEEeqnarray}{rCl}
 \Gamma_{\mathcal{G}_c}(y) & := & \left \| y -  \mathcal{G}_c(y) \right \| \\
\label{eq_150120241844}  & = & \min_{q \in \{1,2,\cdots,Q \}}  \Gamma_{\mathcal{A}_{\mathbf{Y}^{c,q}}}(y).
   \end{IEEEeqnarray}    
\end{definition}
\section{Theory for Learning with Kernel Affine Hull Machines}\label{sec_theory}
This section develops a theory for learning by means of KAHMs. For this, a geometrically inspired kernel function and corresponding RKHS is presented in Section~\ref{sec_160120241607}, followed by a definition of a hypothesis space in Section~\ref{sec_160120241619}. Section~\ref{sec_160120241621} evaluates the Rademacher complexity of the hypothesis space, which is then used in Section~\ref{sec_160120241634} to derive the generalisation error bound for the hypothesis space.     
\subsection{A Novel Kernel Function Induced by the Global KAHM}\label{sec_160120241607}
A given global KAHM, $\mathcal{G}_{c}$, induces a function, $\mathcal{K}_{c}: \mathbb{R}^p \times \mathbb{R}^p \rightarrow [0,1]$, defined as
 \begin{IEEEeqnarray}{rCl}
\label{eq_241220231141}\mathcal{K}_{c}(y^1,y^2) & : = & \exp\left(-\frac{1}{p} \Gamma_{\mathcal{G}_{c}}(y^1)\right) \exp\left(-\frac{1}{p} \Gamma_{\mathcal{G}_{c}}(y^2)\right). 
   \end{IEEEeqnarray} 
$\mathcal{K}_{c}$ is a positive definite kernel, since
\begin{enumerate}
\item $\mathcal{K}_{c}(y^1,y^2) = \mathcal{K}_{c}(y^2,y^1)$, and
\item for every $y^1,\cdots,y^N \in \mathbb{R}^p$ and $\alpha_1,\cdots, \alpha_N \in \mathbb{R}$,  
\begin{IEEEeqnarray}{rCl}
\sum_{i,j=1}^N \alpha_i \alpha_j \mathcal{K}_{c}(y^i,y^j) & \geq & 0. \label{eq_100602112023}
   \end{IEEEeqnarray} 
To see (\ref{eq_100602112023}), consider 
\begin{IEEEeqnarray}{rCl}
\nonumber \sum_{i,j=1}^N \alpha_i \alpha_j \mathcal{K}_{c}(y^i,y^j) & = & \sum_{i,j=1}^N \alpha_i \exp\left(-\frac{1}{p}\Gamma_{\mathcal{G}_{c}}(y^i)\right) \alpha_j \exp\left(-\frac{1}{p}\Gamma_{\mathcal{G}_{c}}(y^j)\right) \\
& = & \left | \sum_{i=1}^N \alpha_i \exp\left(-\frac{1}{p}\Gamma_{\mathcal{G}_{c}}(y^i)\right)  \right |^2 \geq 0.
 \end{IEEEeqnarray} 
\end{enumerate}
\begin{remark}[Justification and Interpretation of $\mathcal{K}_c$]\label{rem_290520241119}
$\mathcal{K}_c(y^1,y^2)$ will be high, only if both $y^1$ and $y^2$ lie close to the $c^{th}$ class labelled samples that may have been owned by any of the $Q$ clients. Thus, $\mathcal{K}_c$ provides a measure of similarity between two data points in-terms of their association to the $c^{th}$ class. That is, $\mathcal{K}_c(y^1,y^2)$ will be high even in the case when $y^1$ and $y^2$ are at far distance from each other but both $y^1$ and $y^2$ are close to some $c^{th}$ class labelled samples. This property of $\mathcal{K}_c$ justifies its choice as kernel function for predicting the association of a point to $c-$th class.      
\end{remark}

Now the RKHS associated to $\mathcal{K}_{c}$ is given by:
\begin{IEEEeqnarray}{rCl}
\nonumber \lefteqn{\mathcal{H}_{\mathcal{K}_{c}}(\mathbb{R}^p)} \\
& : = & \left \{ f = \sum_{i=1}^{\infty} \alpha_i \mathcal{K}_{c}(\cdot,y^i) \: \mid \:  \alpha_i \in \mathbb{R},\; y^i \in \mathbb{R}^p,\;    \| f\|_{\mathcal{H}_{\mathcal{K}_{c}}(\mathbb{R}^p)}^2: = \sum_{i,j=1}^{\infty} \alpha_i \alpha_j \mathcal{K}_{c}(y^i,y^j) < \infty   \right \} \IEEEeqnarraynumspace
 \end{IEEEeqnarray} 
with inner product for any $f(\cdot) = \sum_{i=1}^{L}a_i \mathcal{K}_{c}(\cdot,s^i) $ (with $a^i \in \mathbb{R}, s^i \in \mathbb{R}^p$) and $g(\cdot) = \sum_{j=1}^{M}b_j \mathcal{K}_{c}(\cdot,t^j) \in \mathcal{H}_{\mathcal{K}_{c}}(\mathbb{R}^p)$ (with $b_j \in \mathbb{R}, t^j \in \mathbb{R}^p$) defined as
 \begin{IEEEeqnarray}{rCl}
 \langle f,g\rangle_{\mathcal{H}_{\mathcal{K}_{c}}(\mathbb{R}^p)} & : = & \sum_{i=1}^L \sum_{j=1}^M a_i b_j \mathcal{K}_{c}(s^i,t^j). 
 \end{IEEEeqnarray}
\subsection{A Data-Dependent Hypothesis Space}\label{sec_160120241619}
Let $f_{y \mapsto z_c}:\mathbb{R}^p \to \mathbb{R}$ be a function such that $f_{y \mapsto z_c}(y)$ serves as an approximation to $z_c$, i.e., $f_{y \mapsto z_c}(y)$ predicts the association of the data point $y$ to the $c^{th}$ class. Given the data set $\mathcal{D}$, as defined in (\ref{eq_151220231703}), the hypothesis space for predicting the association of a point to the $c^{th}$ class is defined as a convex hull within $\mathcal{H}_{\mathcal{K}_{c}}(\mathbb{R}^p)$: 
  \begin{IEEEeqnarray}{rCl} 
\label{eq_030120241551}\mathcal{M}_{\mathcal{D},c} & : = & \left \{ f_{y \mapsto z_c} = \sum_{i=1}^{N} \alpha_{c,i} \mathcal{K}_{c}(\cdot,y^i) \; \mid \;  \alpha_{c,i} \in [0,1],\; \sum_{i=1}^{N}   \alpha_{c,i} = 1,\; (y^i,z^i,q^i) \in \mathcal{D}     \right \}. 
  \end{IEEEeqnarray}
It is obvious that
 \begin{IEEEeqnarray}{rCl} 
\mathcal{M}_{\mathcal{D},c}  & \subset & \mathcal{H}_{\mathcal{K}_{c}}(\mathbb{R}^p).
  \end{IEEEeqnarray}
It can be seen that for any $f_{y \mapsto z_c} \in \mathcal{M}_{\mathcal{D},c}$,
\begin{IEEEeqnarray}{rCl}
\label{eq_090420241411}\left \| f_{y \mapsto z_c} \right \|_{\mathcal{H}_{\mathcal{K}_{c}}(\mathbb{R}^p)} & = & \left | \sum_{i =1}^{N} \alpha_{c,i} \exp\left(-\frac{1}{p} \Gamma_{\mathcal{G}_{c}}(y^i)\right)  \right | \\
\label{eq_090420241412} & \leq & 1,
 \end{IEEEeqnarray} 
where (\ref{eq_090420241412}) follows from $\alpha_{c,i} \in [0,1]$ and $\sum_{i =1}^{N} \alpha_{c,i} = 1$. Thus,
\begin{IEEEeqnarray}{rCl}
\label{eq_150120241129}\sup_{f_{y \mapsto z_c} \in \mathcal{M}_{\mathcal{D},c}} \left \| f_{y \mapsto z_c} \right \|_{\mathcal{H}_{\mathcal{K}_{c}}(\mathbb{R}^p)} & \leq & 1.
 \end{IEEEeqnarray}
\subsection{Rademacher Complexity of the Hypothesis Space}\label{sec_160120241621}
To evaluate the Rademacher complexity of the hypothesis space, we introduce $\sigma_1,\cdots,\sigma_N$ as the independent random variables drawn from the Rademacher distribution, and denote $\sigma = (\sigma_1,\cdots,\sigma_N)$. For a given data set $\mathcal{D}$ (defined in (\ref{eq_151220231703})), the empirical Rademacher complexity of the hypothesis space $\mathcal{M}_{\mathcal{D},c}$ is given as
 \begin{IEEEeqnarray}{rCl}
\widehat{\mathcal{R}}_{\mathcal{D}}(\mathcal{M}_{\mathcal{D},c}) & = & \frac{1}{N} \mathop{\mathbb{E}}_{\sigma } \left[ \sup_{f_{y \mapsto z_c} \in \mathcal{M}_{\mathcal{D},c} } \sum_{i=1}^N \sigma_i \: f_{y \mapsto z_c}(y^i) \right]. 
  \end{IEEEeqnarray} 
\begin{theorem}[Bound on the Rademacher Complexity of $\mathcal{M}_{\mathcal{D},c}$]\label{result_rademacher_complexity_hypthesis_space}
Given a dataset $\mathcal{D} \sim (\mathbb{P}_{y,z,q})^N$, as defined in (\ref{eq_151220231703}), we have 
\begin{IEEEeqnarray}{rCl}
\label{eq_150120241133} \widehat{\mathcal{R}}_{\mathcal{D}}(\mathcal{M}_{\mathcal{D},c}) & \leq  & \frac{1}{\sqrt{N}}.
\end{IEEEeqnarray} 
Thus, the expected Rademacher complexity has an upper bound given by 
\begin{IEEEeqnarray}{rCl}
\label{eq_150120241205} \mathop{\mathbb{E}}_{\mathcal{D} \sim (\mathbb{P}_{y,z,q})^N} \left[\widehat{\mathcal{R}}_{\mathcal{D}}(\mathcal{M}_{\mathcal{D},c})\right]& \leq  & \frac{1}{\sqrt{N}}.
\end{IEEEeqnarray} 
  \begin{proof}
The proof is provided in Appendix C. 
 \end{proof}  
\end{theorem}  
\subsection{Generalization Error Bound}\label{sec_160120241634}
We consider the squared loss function and derive generalisation error bound for our hypothesis space. Given a hypothesis $f_{y \mapsto z_c}  \in \mathcal{M}_{\mathcal{D},c}$, let $l_{f_{y \mapsto z_c}}:\mathbb{R}^p \times \{0,1\} \rightarrow \mathbb{R}$ be a loss function defined as
\begin{IEEEeqnarray}{rCl}
\label{eq_100420241905}l_{f_{y \mapsto z_c}}(y,z_c) & : = & |z_c-f_{y \mapsto z_c}(y)|^2. 
\end{IEEEeqnarray} 
Consider the following family of loss functions defined by the hypothesis space $\mathcal{M}_{\mathcal{D},c}$:
\begin{IEEEeqnarray}{rCl}
\mathcal{L}_{\mathcal{D},c} & : = & \left \{l_{f_{y \mapsto z_c}} : (y,z_c) \mapsto |z_c-f_{y \mapsto z_c} (y)|^2 \; \mid \; f_{y \mapsto z_c}  \in \mathcal{M}_{\mathcal{D},c} \right \}.
\end{IEEEeqnarray} 
The empirical Rademacher complexity of $\mathcal{L}_{\mathcal{D},c}$ is given as
\begin{IEEEeqnarray}{rCl}
\widehat{\mathcal{R}}_{\mathcal{D}}(\mathcal{L}_{\mathcal{D},c}) &  = & \frac{1}{N} \mathop{\mathbb{E}}_{\sigma } \left [ \sup_{l_{f_{y \mapsto z_c}} \in \mathcal{L}_{\mathcal{D},c}} \sum_{i=1}^N \sigma_i l_{f_{y \mapsto z_c}}(y^i,(z^i)_c) \right ].
\end{IEEEeqnarray} 
\begin{lemma}[]\label{result_rademacher_complexity_loss_space}
Given a dataset $\mathcal{D} \sim (\mathbb{P}_{y,z,q})^N$, as defined in (\ref{eq_151220231703}), we have 
\begin{IEEEeqnarray}{rCl}
\label{eq_150120241310} \widehat{\mathcal{R}}_{\mathcal{D}}(\mathcal{L}_{\mathcal{D},c}) & \leq  & \frac{2}{\sqrt{N}}, \\
\label{eq_150120241311} \mathop{\mathbb{E}}_{\mathcal{D} \sim (\mathbb{P}_{y,z,q})^N} \left[\widehat{\mathcal{R}}_{\mathcal{D}}(\mathcal{L}_{\mathcal{D},c})\right] & \leq  & \frac{2}{\sqrt{N}}.
\end{IEEEeqnarray}
\begin{proof}
The proof is based on the application of Talagrand's lemma~\cite{MohriRostamizadehTalwalkar18} to get
$\widehat{\mathcal{R}}_{\mathcal{D}}(\mathcal{L}_{\mathcal{D},c})  \leq   2 \widehat{\mathcal{R}}_{\mathcal{D}}(\mathcal{M}_{\mathcal{D},c})$. Now, (\ref{eq_150120241310}) and (\ref{eq_150120241311}) follow immediately from Theorem~\ref{result_rademacher_complexity_hypthesis_space}. For the sake of completeness, a proof starting from scratch is provided in Appendix D.
\end{proof}
\end{lemma}
\begin{theorem}[Data-Dependent Bound on Generalisation Error for Hypothesis Space]\label{result_generalisation_error}
Given a dataset $\mathcal{D} \sim (\mathbb{P}_{y,z,q})^N$ (as defined in (\ref{eq_151220231703})), for any hypothesis $f_{y \mapsto z_c}  \in \mathcal{M}_{\mathcal{D},c}$, the following holds with probability at least $1-\delta$ for any $\delta \in (0,1)$:
\begin{IEEEeqnarray}{rCl}
\mathop{\mathbb{E}}_{(y,z) \sim \mathbb{P}_{y,z}}[l_{f_{y \mapsto z_c}}(y,z_c)]
& \leq & \widehat{\mathbb{E}}_{\mathcal{D}}[l_{f_{y \mapsto z_c}}]  +  \frac{4  }{\sqrt{N}} + \sqrt{\frac{\log(1/\delta)}{2N}},
\end{IEEEeqnarray}
where $l_{f_{y \mapsto z_c}}$ is the loss function (\ref{eq_100420241905}) and $\widehat{\mathbb{E}}_{\mathcal{D}}[l_{f_{y \mapsto z_c}}] $ is the empirical averaged loss value given as
\begin{IEEEeqnarray}{rCl}
\widehat{\mathbb{E}}_{\mathcal{D}}[l_{f_{y \mapsto z_c}}] & = & \frac{1}{N}\sum_{i=1}^N l_{f_{y \mapsto z_c}}(y^i,(z^i)_c). 
\end{IEEEeqnarray} 
\begin{proof}
The proof is provided in Appendix E.
\end{proof}
\end{theorem}
\begin{remark}[Choice of the Loss Function]\label{remark_290520241602}
We have considered the squared loss function for our analysis, nevertheless, the analysis can be extended to any $\rho-$Lipschitz loss function.
\end{remark}
\section{KAHMs Based Collaborative Learning}\label{sec_210420241350}
This section shows how the global model learning problem can be formulated mathematically  (cf. Section \ref{sec_learning_problem_formulation}) and solved (cf. Section~\ref{sec_120420241334}) to derive a predictor. In addition, theoretical guarantees on the performance of the proposed predictor are provided in Section~\ref{sec_120420241936}. The obtained theoretical results are applied in Section~\ref{sec_classification_application} to solve the classification problem in a federated setting. 

\subsection{Global Model Learning Problem}\label{sec_learning_problem_formulation}
The model learning problem consists of determining a function $f_{y \mapsto z_c} \in \mathcal{M}_{\mathcal{D},c}$, such that the  generalisation error over that function, written $\mathop{\mathbb{E}}_{(y,z) \sim \mathbb{P}_{y,z}}[l_{f_{y \mapsto z_c}}(y,(z)_c)]$, is as small as possible. Since $f_{y \mapsto z_c} \in \mathcal{M}_{\mathcal{D},c}$, we have $f_{y \mapsto z_c}(y)  = \sum_{i=1}^{N} \alpha_{c,i}\mathcal{K}_{c}(y,y^i)$, with $\alpha_{c,i} \in [0,1]$ and $\sum_{i=1}^{N}  \alpha_{c,i} = 1$. The value $\mathcal{K}_{c}(y,y^i)$ will be close to 1 if and only if $\Gamma_{\mathcal{G}_{c}}(y) \approx 0$ and $\Gamma_{\mathcal{G}_{c}}(y^i) \approx 0$ (i.e. if and only if both $y$ and $y^i$ lie close to the $c^{th}$ class labelled training data samples). Similarly $\mathcal{K}_{c}(y,y^i)$ will be close to 0 if either or both $y$ and $y^i$ lie far away from the $c^{th}$ class labelled training data samples. Based on the observation that the value $\mathcal{K}_{c}(y,y^i)$ is the degree of similarity between $y$ and $y^i$ in terms of their distance from the $c^{th}$ class labelled training data samples, $\alpha_{c,i}$ (which is the weight assigned to $\mathcal{K}_{c}(y,y^i)$ in estimating $z_c$) can be chosen as 
\begin{IEEEeqnarray}{rCl}
\label{eq_150120241734} \alpha_{c,i} & = & 0,\; i \notin \mathop{\cup}_{q=1}^Q \mathcal{I}^{c,q},
\end{IEEEeqnarray}
where $\mathcal{I}^{c,q}$ (as defined in (\ref{eq_040120241522})) is the set of indices of the samples which are $c^{th}$ class labelled and owned by $q^{th}$ party. Eq. (\ref{eq_150120241734}) implies that the {\em weight assigned to a non $c^{th}$ class labelled training data sample in estimating $z_c$ is zero}. As a result of (\ref{eq_150120241734}), our learning space (within the hypothesis space $\mathcal{M}_{\mathcal{D},c}$) is given as  
  \begin{IEEEeqnarray}{rCl} 
\nonumber \lefteqn{\widetilde{\mathcal{M}}_{\mathcal{D},c} } \\
& : = & \left \{ f_{y \mapsto z_c} = \sum_{q=1}^Q \sum_{i=\mathrm{I}_1^{c,q}}^{\mathrm{I}_{|\mathcal{I}^{c,q}|}^{c,q}} \alpha_{c,i} \mathcal{K}_{c}(\cdot,y^i) \; \mid \;  \alpha_{c,i} \in [0,1],\; \sum_{q=1}^Q \sum_{i=\mathrm{I}_1^{c,q}}^{\mathrm{I}_{|\mathcal{I}^{c,q}|}^{c,q}}   \alpha_{c,i} = 1,\; (y^i,z^i,q^i) \in \mathcal{D}     \right \} \IEEEeqnarraynumspace \\
& \subset & \mathcal{M}_{\mathcal{D},c}.
  \end{IEEEeqnarray}
With $\widetilde{\mathcal{M}}_{\mathcal{D},c}$ as the learning space, the learning problem can be formulated as 
\begin{problem}[Learning Problem]\label{problem_learning}
Given a dataset $\mathcal{D} \sim (\mathbb{P}_{y,z,q})^N$ (as defined in (\ref{eq_151220231703})), the learning problem is formulated as
  \begin{IEEEeqnarray}{rCl} 
  f_{y \mapsto z_c}^* & = &  \mathop{\argmin}_{  f_{y \mapsto z_c} \in \widetilde{\mathcal{M}}_{\mathcal{D},c}} \; \mathop{\mathbb{E}}_{(y,z) \sim \mathbb{P}_{y,z}}\left [ \left |z_c- f_{y \mapsto z_c}(y)\right |^2 \right ].
    \end{IEEEeqnarray}
\end{problem}
\subsection{A Solution of the Learning Problem}\label{sec_120420241334}
The solution of Problem~\ref{problem_learning} is challenging without making any assumptions about the unknown distribution $\mathbb{P}_{y,z,q}$. However, fortunately a workaround exists for approximating $f_{y \mapsto z_c}^*$ without directly solving Problem~\ref{problem_learning}. For this, a {\em realistic assumption} is made:
\begin{assumption}[]\label{assumption_1}
The training data samples of $c^{th}$ class (i.e. $\{y^i \; \mid \; i \in \mathop{\cup}_{q=1}^Q \mathcal{I}^{c,q}  \}$) are fitted by the global KAHM $\mathcal{G}_c$ with sufficient accuracy such that
 \begin{IEEEeqnarray}{rCl}
\exp\left(- \frac{\Gamma_{\mathcal{G}_c}(y^i) }{p} \right)
& \approx & 1,\; \forall i \in \mathop{\cup}_{q=1}^Q \mathcal{I}^{c,q}.
  \end{IEEEeqnarray} 
\end{assumption}
Assumption~\ref{assumption_1} requires that $y^i \approx \mathcal{G}_c(y^i)$, i.e. $\Gamma_{\mathcal{G}_c}(y^i) \approx 0$, $\forall i \in \mathop{\cup}_{q=1}^Q \mathcal{I}^{c,q}$. In other words, the fitting error on the training data samples by KAHM should be small, which is a realistic assumption. This assumption will be also validated through experiments later on.
\begin{theorem}[An Approximate Solution to the Learning Problem]\label{theorem_learning_solution}
Given a dataset $\mathcal{D} \sim (\mathbb{P}_{y,z,q})^N$ (as defined in (\ref{eq_151220231703})), the solution of Problem~\ref{problem_learning}, under Assumption~\ref{assumption_1}, is given by
  \begin{IEEEeqnarray}{rCl}
\label{eq_150120241814} f_{y \mapsto z_c}^*(y)&\approx& \exp\left(-\frac{1}{p} \Gamma_{\mathcal{G}_c}(y)\right).
   \end{IEEEeqnarray}  
   \begin{proof}
The proof is provided in Appendix F.
   \end{proof}
\end{theorem}
\begin{remark}[Significance of Learning solution]\label{rem_significance_approximate_solution}
The significance of Theorem~\ref{theorem_learning_solution}'s  approximate solution lies in the fact that $f_{y \mapsto z_c}^*(y)$ can be evaluated without estimating any of the model parameters, leading to an efficient solution that does not require a gradient-based learning algorithm.
\end{remark}
\subsection{Theoretical Guarantees}\label{sec_120420241936}
Since $f_{y \mapsto z_c}^* \in \mathcal{M}_{\mathcal{D},c}$, an upper bound on the generalisation error of $f_{y \mapsto z_c}^*$ is provided by Theorem~\ref{result_generalisation_error}. However, a more tight bound can be obtained in the light of Assumption~\ref{assumption_1} and by making another reasonable assumption:
\begin{assumption}[]\label{assumption_2}
The non $c^{th}$ class training data samples (that is, $\{y^i \; \mid \; i \notin \mathop{\cup}_{q=1}^Q \mathcal{I}^{c,q}  \}$) are {not well fitted} by the $c^{th}$ class associated global KAHM $\mathcal{G}_c$ such that
 \begin{IEEEeqnarray}{rCl}
\exp\left(- \frac{\Gamma_{\mathcal{G}_c}(y^i) }{p} \right)
& \approx & 0,\; \forall i \notin \mathop{\cup}_{q=1}^Q \mathcal{I}^{c,q}.
  \end{IEEEeqnarray} 
\end{assumption}
Assumption \ref{assumption_2} is practical, since class $c$'s associated global KAHM $\mathcal{G}_c$ has been learned to fit only the class $c$ labelled samples, and thus non-$c$ class labelled training samples can not be reconstructed using $\mathcal{G}_c$, resulting in a large value of $\Gamma_{\mathcal{G}_c}$ at non-$c$ class labelled training samples. 
\begin{theorem}[Bound on Generalization Error of Predictor $f_{y \mapsto z_c}^*$]\label{result_generalisation_error_2}
Given a dataset $\mathcal{D} \sim (\mathbb{P}_{y,z,q})^N$ (as defined in (\ref{eq_151220231703})), for the predictor $f_{y \mapsto z_c}^*$ (as defined in (\ref{eq_150120241814})), under Assumption~\ref{assumption_1} and Assumption~\ref{assumption_2}, the following holds with probability at least $1-\delta$ for any $\delta \in (0,1)$:
\begin{IEEEeqnarray}{rCl}
\mathop{\mathbb{E}}_{(y,z) \sim \mathbb{P}_{y,z}}\left [\left |z_c-f_{y \mapsto z_c}^*(y) \right|^2 \right ]
& \leq &  \frac{4  }{\sqrt{N}} + \sqrt{\frac{\log(1/\delta)}{2N}}.
\end{IEEEeqnarray}
\begin{proof}
The proof is provided in Appendix G.
\end{proof} 
\end{theorem}
Since we are considering the multi-class classification problem, an analysis of the error in approximating the class label probability (conditioned on a given input sample) is of interest. We are interested in upper bounding the mismatch between predictor $f_{y \mapsto z_c}^*(y)$ and the target function $\mathbb{P}_{z | y}(z_c = 1 | y)$.   
\begin{theorem}[Target Function Approximation Error Bound]\label{result_generalisation_error_3}
Given a dataset $\mathcal{D} \sim (\mathbb{P}_{y,z,q})^N$ (as defined in (\ref{eq_151220231703})), for the predictor $f_{y \mapsto z_c}^*$, 
\begin{itemize}
\item  the following holds with probability at least $1-\delta$ for any $\delta \in (0,1)$: 
\begin{IEEEeqnarray}{rCl}
\nonumber \mathop{\mathbb{E}}_{y \sim \mathbb{P}_{y}}\left [ |f_{y \mapsto z_c}^*(y) -  \mathbb{P}_{z | y}(z_c = 1 | y) |^2\right]  & \leq & \frac{1}{N}\sum_{i=1}^N \left |(z^i)_c-f_{y \mapsto z_c}^*(y^i) \right|^2 \\
 & &  +  \frac{4  }{\sqrt{N}} + \sqrt{\frac{\log(1/\delta)}{2N}}. 
\end{IEEEeqnarray} 
\item under Assumption~\ref{assumption_1} and Assumption~\ref{assumption_2}, the following holds with probability at least $1-\delta$ for any $\delta \in (0,1)$:
\begin{IEEEeqnarray}{rCl}
 \mathop{\mathbb{E}}_{y \sim \mathbb{P}_{y}}\left [ |f_{y \mapsto z_c}^*(y) -  \mathbb{P}_{z | y}(z_c = 1 | y) |^2\right] 
& \leq &  \frac{4  }{\sqrt{N}} + \sqrt{\frac{\log(1/\delta)}{2N}}. 
\end{IEEEeqnarray} 
\end{itemize}
\begin{proof}
The proof is provided in Appendix H. 
\end{proof}
\end{theorem}
\begin{lemma}[Practical Significance of Theorem~\ref{result_generalisation_error_3}]\label{lemma_300420240936}
The practical significance of Theorem~\ref{result_generalisation_error_3} is for approximating the target function. Theorem~\ref{result_generalisation_error_3} allows us to make the following assumption: 
\begin{IEEEeqnarray}{rCl}
\label{eq_300420240945}f_{y \mapsto z_c}^*(y) & \approx & \mathbb{P}_{z | y}(z_c = 1 | y \sim \mathbb{P}_{y} ).
\end{IEEEeqnarray}
Using (\ref{eq_150120241814}) and (\ref{eq_150120241844}) in (\ref{eq_300420240945}), we get
\begin{IEEEeqnarray}{rCl}
\label{eq_170420241410}\min_{q \in \{1,2,\cdots,Q \}}  \Gamma_{\mathcal{A}_{\mathbf{Y}^{c,q}}}(y) & \approx & -p \log\left(\mathbb{P}_{z | y}(z_c = 1 | y \sim \mathbb{P}_{y} ) \right).
\end{IEEEeqnarray}
\end{lemma}
Theorem~\ref{result_generalisation_error_3} further allows us to determine the sample complexity for $f_{y \mapsto z_c}^*$, as stated in the Lemma~\ref{lemma_result_generalisation_error_4}:  
\begin{lemma}[Sample Complexity for $f_{y \mapsto z_c}^*$]\label{lemma_result_generalisation_error_4}
The number of data points, needed to be sampled from a distribution $\mathbb{P}_{y,z,q}$ to guarantee
\begin{IEEEeqnarray}{rCl}
\mathop{\mathbb{E}}_{y \sim \mathbb{P}_{y}}\left [ |f_{y \mapsto z_c}^*(y) -  \mathbb{P}_{z | y}(z_c = 1 | y) |^2\right]  & \leq & \epsilon
\end{IEEEeqnarray} 
with probability at least $1-\delta$ for any $\delta \in (0,1)$ and $\epsilon > 0$ under Assumption~\ref{assumption_1} and Assumption~\ref{assumption_2}, is given as 
\begin{IEEEeqnarray}{rCl}
N(\epsilon,\delta) & = & \Omega \left( \frac{1}{\epsilon^2} \left(4 + \sqrt{\frac{\log(1/\delta)}{2}} \right)^2 \right).
\end{IEEEeqnarray} 
\begin{figure}  
\centering
\includegraphics[width=0.7\textwidth]{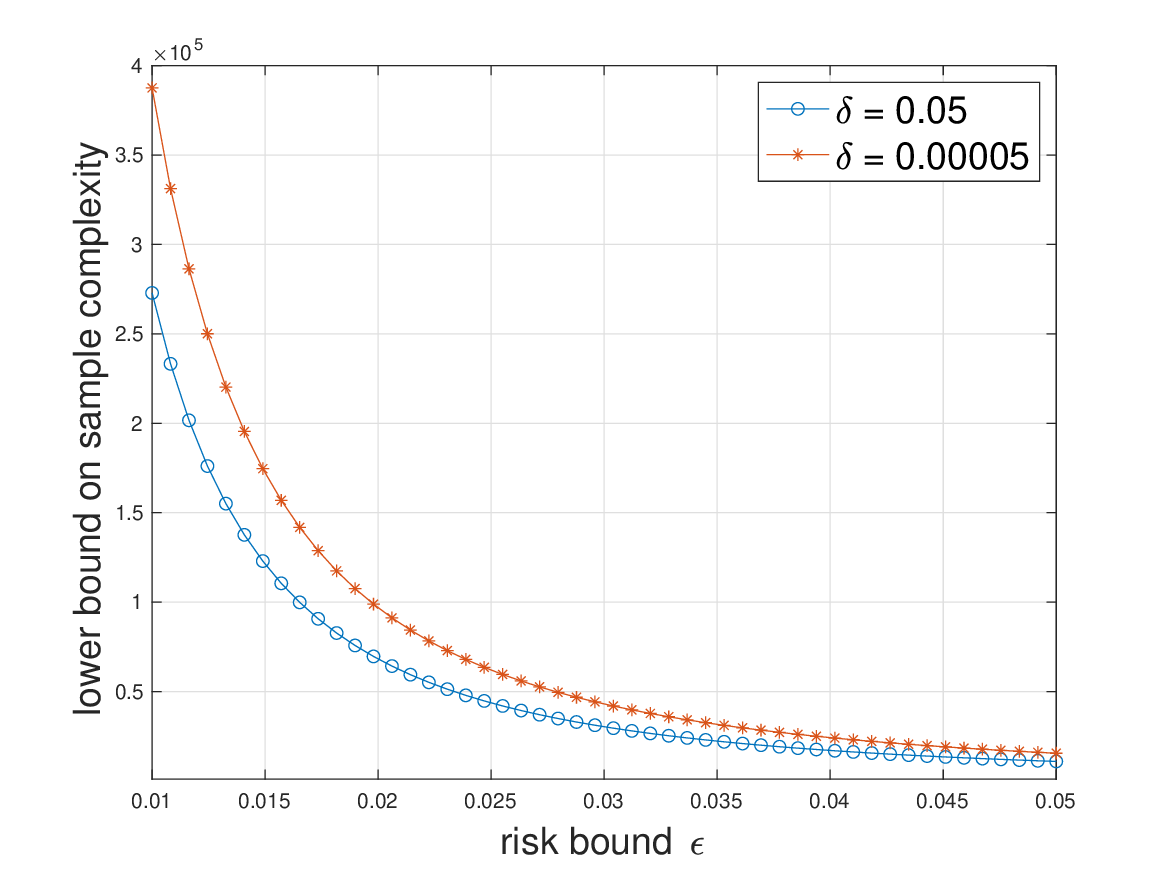}\label{fig_sample_complexity}
\caption{A plot of the lower bound on sample complexity against target function approximation risk bound.}
\end{figure}
Figure~\ref{fig_sample_complexity} plots the lower bound on sample complexity against target function approximation risk bound for two different values of failure probabilities. 
\end{lemma}
\begin{theorem}[Deterministic Analysis of Predictor $f_{y \mapsto z_c}^*$]\label{theorem_deterministic_analysis}
The degree of membership of a point $y \in \mathbb{R}^p$ to the $c^{th}$ class, assigned by the predictor $f_{y \mapsto z_c}^*$ under Assumption~\ref{assumption_1}, is related to the distance of that point from the $c^{th}$ class labelled training samples as in the following:      
\begin{IEEEeqnarray}{rCl}
\nonumber \lefteqn{f_{y \mapsto z_c}^*(y)}\\
\label{eq_030520242103}& > & \exp \left(- \left(\frac{1}{p} + \frac{ \left |\mathcal{I}^{c,q^*(y)} \right |^2 }{2 \left\|\mathbf{Y}^{\displaystyle c,q^*(y) } \right \|^2_F} \right) \left \| \left[\begin{IEEEeqnarraybox*}[][c]{,c/c/c,} y - y^{\displaystyle \mathrm{I}_1^{c,q^*(y)}} & \cdots & y - y^{\displaystyle \mathrm{I}_{\left|\mathcal{I}^{c,q^*(y)}\right|}^{c,q^*(y)}} \end{IEEEeqnarraybox*} \right]   \right \|_2 \right),
\end{IEEEeqnarray}
where
\begin{IEEEeqnarray}{rCl}
q^*(y) & = & \mathop{\argmin}_{q \in \{1,2,\cdots,Q \}}\; \Gamma_{\mathcal{A}_{\mathbf{Y}^{c,q}}}(y). 
\end{IEEEeqnarray} 
\begin{proof}
The proof is provided in Appendix I.
\end{proof}
\end{theorem}
\begin{remark}[Significance of Theorem~\ref{theorem_deterministic_analysis}]\label{rem_070520241636}
Theorem~\ref{theorem_deterministic_analysis} implies that if a point $y$ is close to the $c^{th}$ class labelled and $q^*(y)-$th client owned training samples $\left\{y^{\displaystyle \mathrm{I}_1^{c,q^*(y)}},\cdots,y^{\displaystyle \mathrm{I}_{\left|\mathcal{I}^{c,q^*(y)}\right|}^{c,q^*(y)}}\right\}$ (i.e. $\left \| \left[\begin{IEEEeqnarraybox*}[][c]{,c/c/c,} y - y^{\displaystyle \mathrm{I}_1^{c,q^*(y)}} & \cdots & y - y^{\displaystyle \mathrm{I}_{\left|\mathcal{I}^{c,q^*(y)}\right|}^{c,q^*(y)}} \end{IEEEeqnarraybox*} \right]   \right \|_2$ is small), then the value $f_{y \mapsto z_c}^*(y)$ cannot be small. Thus, a small value of $f_{y \mapsto z_c}^*(y)$ indicates that $y$ is at a far distance from the $c^{th}$ class labelled training samples of all clients, and thus an interpretation of the predictor can be given in terms of the distance from training samples.
\end{remark}
\subsection{Classification Applications in Federated Setting}\label{sec_classification_application}   
\begin{definition}[A Global Classifier]\label{def_global_classifier}
Given a distributed dataset $\mathcal{D} \sim (\mathbb{P}_{y,z,q})^N$ (as defined in (\ref{eq_151220231703})), a global classifier, $\mathcal{C}:\mathbb{R}^p \rightarrow \{1,2,\cdots,C\}$, is defined as
\begin{IEEEeqnarray}{rCl}
\mathcal{C}(y) & =  &\mathop{\argmax}_{c \in \{1,2,\cdots,C\}} \mathbb{P}_{z | y}(z_c = 1 | y \sim \mathbb{P}_{y} ).
   \end{IEEEeqnarray} 
Using (\ref{eq_170420241410}), $\mathcal{C}(y)$ can be approximated as
\begin{IEEEeqnarray}{rCl}
\label{eq_150120241845}\widehat{\mathcal{C}}(y) & =  & \mathop{\argmin}_{c \in \{1,2,\cdots,C\}} \left( \min_{q \in \{1,2,\cdots,Q \}}  \Gamma_{\mathcal{A}_{\mathbf{Y}^{c,q}}}(y) \right). 
   \end{IEEEeqnarray}  
   \end{definition}
A local classifier is derived from the global classifier by staying confined to the local data as in Definition~\ref{def_local_classifier}: 
\begin{definition}[A Local Classifier]\label{def_local_classifier}
For the $q-$th client with data $\{y^i \; \mid \; i \in \cup_{c=1}^C\mathcal{I}^{c,Q} \}$, the local classifier, $\widehat{\mathcal{C}}_q: \mathbb{R}^p \rightarrow \{1,2,\cdots,C\}$, is defined as
\begin{IEEEeqnarray}{rCl}
\label{eq_170420241923}\widehat{\mathcal{C}}_q(y) & =  & \mathop{\argmin}_{c \in \{1,2,\cdots,C\}} \Gamma_{\mathcal{A}_{\mathbf{Y}^{c,q}}}(y). 
   \end{IEEEeqnarray}        
\end{definition}
\begin{remark}[Local Data with Missing Classes]\label{rem_300420241017}
If the $q-$th client has zero $c^{th}$ class labelled samples, then (\ref{eq_150120241845}) is evaluated taking $\Gamma_{\mathcal{A}_{\mathbf{Y}^{c,q}}}(y) = \infty$. 
\end{remark}
\begin{remark}[Addressing Computational Challenge of Big Datasets] \label{remark_wide_form}
The computational challenge of KAHM modelling of big datasets can be addressed, as suggested by~\cite{KAHM}, by partitioning the total dataset into subsets and modelling each subset through a separate KAHM. Specifically, if $ |\mathcal{I}^{c,q}|$ (i.e. the number of $c^{th}$ class labelled samples that are owned by client $q$, that equals the number of rows in matrix $\mathbf{Y}^{c,q}$) is more than 1000, then we define
\begin{IEEEeqnarray}{rCl}
S & = & \lceil  |\mathcal{I}^{c,q}| / 1000 \rceil \\
\{ \mathbf{Y}^{c,q}_1,\cdots, \mathbf{Y}^{c,q}_{S} \} & = & \clustering\left(\{(\mathbf{Y}^{c,q})_{1,:},\cdots, (\mathbf{Y}^{c,q})_{ |\mathcal{I}^{c,q}|,:} \},S\right) \\
\label{eq_220420241354}\Gamma_{\mathcal{A}_{\mathbf{Y}^{c,q}}}(y) & := & \min_{s \in \{1,2,\cdots,S\}}\; \Gamma_{\mathcal{A}_{\mathbf{Y}^{c,q}_s}}(y).
   \end{IEEEeqnarray}  
That is, the total data samples (stored in the rows of matrix $\mathbf{Y}^{c,q}$) are partitioned into a number of subsets $S$ (where $S$ equals the rounding of $|\mathcal{I}^{c,q}|/1000$ towards the nearest integer) through clustering, and each subset is modelled through a separate KAHM resulting in a set of KAHMs $\{\mathcal{A}_{\mathbf{Y}^{c,q}_1},\cdots,\mathcal{A}_{\mathbf{Y}^{c,q}_S} \}$, which are finally aggregated through the distance measure given in (\ref{eq_220420241354}).             
\end{remark}
\begin{remark}[Practical Significance of Global Classifier (\ref{eq_150120241845})]\label{rem_300420240952}
The significance of the global classifier (\ref{eq_150120241845}) is that its evaluation does not require individual KAHMs (that are owned by different clients), but only  the distance measures, giving rise to a collaborative learning scheme as shown in Figure~\ref{fig_KAHM_distributed_classifier}. Concretely, the collaborative learning scheme sketched in Figure~\ref{fig_KAHM_distributed_classifier} requires a processing of user inputs by client's local models. The passing of user inputs to an arbitrary client for local inference can be avoided by transferring all of the local KAHMs $\{\{\mathcal{A}_{\mathbf{Y}^{c,q}}\}_{c=1}^C\}_{q=1}^Q$ to the server. 
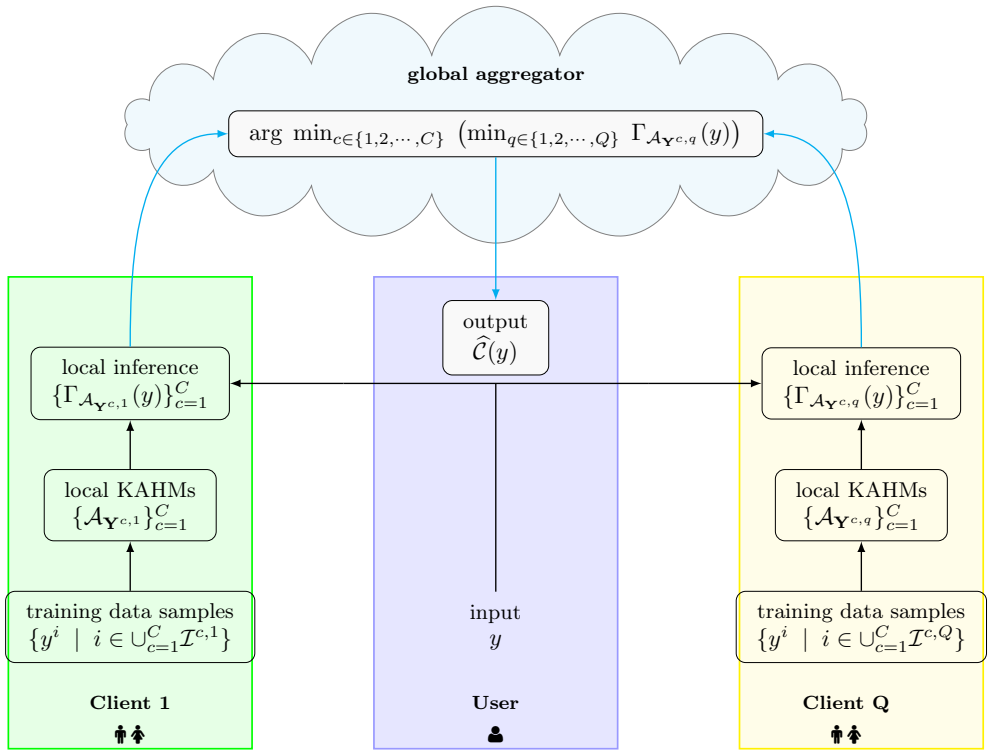
\begin{figure}[!h]  
\centering
\scalebox{0.9}{\begin{tikzpicture}[scale=0.9]
\path[fill=green!10](-2,-8.25)--(2,-8.25)--(2,-0.5)--(-2,-0.5)--cycle;
\draw[green,line width = 0.25mm](-2,-8.25)--(2,-8.25)--(2,-0.5)--(-2,-0.5)--cycle;
\draw (0,-7.75) node[]{\bfseries $\begin{array}{c} \mbox{\scriptsize Client 1} \\ \mbox{\scriptsize \faMale\:\faFemale} \end{array}$};
\draw (0,-6.25) node[rounded corners,draw](n1){\footnotesize $\begin{array}{c}\mbox{training data samples}  \\ \mbox{\small $\{y^i \; \mid \; i \in \cup_{c=1}^C \mathcal{I}^{c,1} \}$} \end{array}$};
\draw (0,-4.25) node[rounded corners,draw](nadd1){\footnotesize $\begin{array}{c}\mbox{local KAHMs}  \\ \mbox{\small $\{ \mathcal{A}_{\mathbf{Y}^{c,1}}\}_{c=1}^C$}\end{array}$};
\draw[-latex,line width=0.2mm] (n1) to [out=90,in=-90] (nadd1);  
\draw (0,-2.25) node[rounded corners,draw](n4){ \footnotesize $\begin{array}{c}\mbox{local inference} \\ \mbox{\small $\{  \Gamma_{\mathcal{A}_{\mathbf{Y}^{c,1}}}(y) \}_{c=1}^C$} \end{array}$};
\draw[-latex,line width=0.2mm] (nadd1) to [out=90,in=-90] (n4);
\path[fill=yellow!10](10,-8.25)--(14,-8.25)--(14,-0.5)--(10,-0.5)--cycle;
\draw[yellow,line width = 0.25mm](10,-8.25)--(14,-8.25)--(14,-0.5)--(10,-0.5)--cycle;
\draw (11.75,-7.75) node[]{\bfseries $\begin{array}{c} \mbox{\scriptsize Client Q} \\ \mbox{\scriptsize \faMale\:\faFemale} \end{array}$};
\draw (12,-2.25) node[rounded corners,draw](n11){ \footnotesize $\begin{array}{c}\mbox{local inference} \\ \mbox{\small $\{  \Gamma_{\mathcal{A}_{\mathbf{Y}^{c,q}}}(y) \}_{c=1}^C$} \end{array}$};
\draw (12,-4.25) node[rounded corners,draw](nadd2){\footnotesize $\begin{array}{c}\mbox{local KAHMs}  \\ \mbox{\small $\{ \mathcal{A}_{\mathbf{Y}^{c,q}}\}_{c=1}^C$}\end{array}$};
\draw[-latex,line width=0.2mm] (nadd2) to [out=90,in=-90] (n11);  
\draw (12,-6.25) node[rounded corners,draw](n15){\footnotesize $\begin{array}{c}\mbox{training data samples}  \\ \mbox{\small $\{y^i \; \mid \; i \in \cup_{c=1}^C\mathcal{I}^{c,Q} \}$} \end{array}$};
\draw[-latex,line width=0.2mm] (n15) to [out=90,in=-90] (nadd2);  
\path[fill=blue!10](4,-8.25)--(8,-8.25)--(8,-0.5)--(4,-0.5)--cycle;   
\draw[blue!40,line width = 0.25mm](4,-8.25)--(8,-8.25)--(8,-0.5)--(4,-0.5)--cycle; 
\draw (6,-7.75) node[]{\bfseries $\begin{array}{c} \mbox{\scriptsize User} \\ \mbox{\scriptsize \faUser} \end{array}$};
  \draw (6,-6.25) node[](n20){ \footnotesize $\begin{array}{c}\mbox{input} \\ \mbox{\small $y$} \end{array}$};
\draw[thick,line width=0.2mm](n20) -- (6,-2.25);
\draw[thick,line width=0.2mm](3.5,-2.25) -- (8.5,-2.25);
\draw[-latex,thick,line width=0.2mm] (3.5,-2.25) to [out=180,in=0] (n4);   
 \draw[-latex,thick,line width=0.2mm] (8.5,-2.25) to [out=0,in=180] (n11); 
  \node[cloud,
    draw = gray,
    fill = cyan!5,
    minimum width = 11cm,
    minimum height = 3.5cm,
    cloud puffs = 18] (c) at (6,2) {};
\draw (6,1.85) node[rounded corners,draw, fill=gray!5](n8){ \footnotesize $\begin{array}{c} \mbox{\small $ \arg \; \min_{c \in \{1,2,\cdots,C \}}\; \left( \min_{q \in \{1,2,\cdots,Q \}}\;\Gamma_{\mathcal{A}_{\mathbf{Y}^{c,q}}}(y)\right)$}  \end{array}$};
\draw[-latex,line width=0.2mm,cyan] (n4) to [out=90,in=180] (n8);  
\draw (6,2.8) node[]{\bfseries {\scriptsize $\begin{array}{c} \mbox{global aggregator} \end{array}$  }};
\draw[-latex,line width=0.2mm,cyan] (n11) to [out=90,in=0] (n8);  
\draw (6,-1.5) node[rounded corners,draw, fill = gray!5](n17){ \footnotesize $\begin{array}{c}\mbox{output} \\ \mbox{\small $\widehat{\mathcal{C}}(y) $} \end{array}$};
\draw[-latex,line width=0.2mm,cyan] (n8) to [out=-90,in=90] (n17);  
\end{tikzpicture}}
\caption{A representation of the collaborative learning solution. For practical implementation, the local KAHMs $\{\{\mathcal{A}_{\mathbf{Y}^{c,q}}\}_{c=1}^C\}_{q=1}^Q$ can be transferred to the server to avoid the passing of user's input to any client for local inference. {\em The solution of global model learning problem has been derived without making any statistical assumptions regarding clients' data distributions, thereby providing by design a robustness towards statistical heterogeneity}.}
\label{fig_KAHM_distributed_classifier}
\end{figure}
\end{remark}
\section{Experiments}\label{sec_210420241503}
Existing research~\cite{KAHM,kumar2023secure} has shown not only the privacy-preserving potential of KAHMs (by fabricating privacy-preserving data) but also that they remain computationally practical, given that they are capable of addressing computational challenges of big data (as stated in Remark~\ref{remark_wide_form}). Thus, experiments to establish the privacy-preserving property and computational efficiency are not repeated here. However, Assumption~\ref{assumption_1} made here to derive a learning solution (cf. Theorem~\ref{theorem_learning_solution}) still needs to be validated through experiments. Consequently, we have conducted experiments to this effect (see Section~\ref{sec_210420241513}). Further, the competitive advantage in terms of performance of a KAHM-based approach to collaborative learning in a federated setting still needs to be established by comparing it to the state of the art methods. Federated learning experiments are provided in Section~\ref{sec_240420241304}, followed by experiments of knowledge transfer across clients in Section~\ref{sec_260420241120}. Finally, in Section~\ref{sec_120520242021}, we show the effectiveness of our method in the single-class data scenario.
\paragraph{Implementation.} The method was implemented using MATLAB (R2024a) and the source code was made publicly available on \url{https://github.com/software-competence-center-hagenberg/GIKM}. The experiments were performed on an iMac (M1, 2021) machine with 8 GB RAM. It is worth mentioning that the proposed method does not involve any free parameters to be tuned, since building a KAHM requires only selecting the subspace dimension $n \leq p$, which is determined as stated in Appendix B. The computational challenge of big datasets is addressed as stated in Remark~\ref{remark_wide_form}. Having defined client and class specific KAHMs (in accordance to Definition~\ref{def_affine_hull_model}) from the available training data samples, no additional computational algorithm is required for the inference of both global (\ref{eq_150120241845}) and local classifiers (\ref{eq_170420241923}). 
\paragraph{Datasets.} We study the multi-class classification problem on the benchmark datasets including MNIST, Freiburg Groceries, Fashion MNIST, CIFAR-10, CIFAR-100, and Office-Caltech-10 datasets. The MNIST dataset contains $28 \times 28$ sized images (of digits) divided into 60000 training images and 10000 test images. The Freiburg Groceries dataset~\shortcite{DBLP:journals/corr/JundAEB16} has 4947 images of grocery products (commonly sold in Germany) labelled into 25 different categories and divided into 3929 training images and 1018 test images. The Fashion MNIST dataset contains 60000 training and 10000 test 28$\times$28 grayscale images of fashion products from 10 categories. The CIFAR-10 dataset contains 50000 training and 10000 test 32$\times$32 color images from 10 different classes. The CIFAR-100 dataset consists of 60000 32$\times$32 color images from 100 classes with 500 training images and 100 test images in each class. The Office-Caltech-10 dataset, containing the 10 overlapping categories between the Office dataset and Caltech256 dataset, consists of images coming from four data sources: Amazon (958 images), Caltech (1123 images), DSLR (157 images), and Webcam (295 images).    
\paragraph{Data Processing.} KAHMs are built from training data samples. Since our experiments are on the images, a feature vector needs to be extracted from each image so that the client and class specific KAHMs could be built from the extracted feature vectors. For MNIST and Fashion MNIST datasets, $28 \times 28$ pixel values of each image are divided by 255 (to scale the values in the range from 0 to 1) and flattened to an equivalent $784-$dimensional data point. For Freiburg Groceries, CIFAR-10, CIFAR-100, and Office-Caltech-10 datasets, the existing ResNet-50 neural network is employed as feature extractor by using the activations of ``avg\_pool'' layer (i.e. the last average pooling layer just before the fully connected layer) as features, resulting into a $2048-$dimensional data point for each image. For Office-Caltech10 images in transfer learning experiments, additionally a 4096-dimensional data point is computed from the activations of ``fc6'' layer of the existing VGG-16 neural network to compare the results with previous studies using same features. Finally, for all the datasets, the hyperbolic tangent function operates along each dimension of a data point to constrain the values between -1 and +1, resulting in the feature vectors to be considered for classification. 
\subsection{Validation of the Underlying Assumption for the Proposed Solution}\label{sec_210420241513} 
To check the validity of Assumption~\ref{assumption_1} in the experiments, the following score is defined:
\begin{IEEEeqnarray*}{rCl}
E  & = & \max_{\displaystyle c \in \{1,\cdots,C\}} \max_{\displaystyle i \in \mathop{\cup}_{q=1}^Q \mathcal{I}^{c,q}} \left| 1 - \exp\left(- \frac{\Gamma_{\mathcal{G}_c}(y^i) }{p} \right) \right|. 
   \end{IEEEeqnarray*} 
\begin{table}
    \centering
    \begin{tabular}{c|cccc}
        \toprule
        \textbf{dataset}   & MNIST & Freiburg Groceries & CIFAR-10 & CIFAR-100  \\
        \midrule
        \textbf{E}  & 0.0045  & 0.0039   & 0.0055 &  0.0049   \\ \midrule
        \textbf{Accuracy} & 0.9870 & 0.8969 &  0.9122 & 0.7471 \\
                \bottomrule
    \end{tabular}
    \caption{Validation of Assumption~\ref{assumption_1} on MNIST, Freiburg Groceries, CIFAR-10, and CIFAR-100 datasets.}
    \label{tab:1}
\end{table}
A low value of $E$ (close to zero) will indicate the validity of Assumption~\ref{assumption_1}, whereas if the converse is true our assumption does not hold. A single-class data scenario has been created assuming that training data samples of a class are completely owned by a single client. Thus, the number of clients is equal to the number of classes. The observed $E$ value and accuracy obtained by the global classifier (\ref{eq_150120241845}) for the different datasets are provided in Table~\ref{tab:1}. The low values of $E$, with $E< 0.01$, across all of the considered datasets validate Assumption~\ref{assumption_1}.
\subsection{Federated Learning}\label{sec_240420241304}
Following~\cite{10.1609/aaai.v37i8.26197}, we consider a non-iid label skew 20\% (or 30\%) federated learning setting, in which
\begin{itemize}
\item the number of clients is equal to 100;
\item each client is first randomly assigned 20\% (or 30\%) of the total available class-labels in a dataset, and then the training samples of each class are randomly distributed equally among clients who have been assigned that class;
\item all of the test samples of a class are assigned to every client who has been assigned that class; 
\item the accuracy over the client's test data, averaged across clients, is calculated to evaluate the performance.      
\end{itemize}
\begin{table}
\centering
\begin{tabular}{cccc}
\toprule
 Method & Fashion MNIST & CIFAR-10 & CIFAR-100 \\
 \midrule
 \texttt{KAHM} Global Classifier (\ref{eq_150120241845}) & \bfseries 95.24 & \bfseries 96.32 & \bfseries 82.4 \\
  \texttt{FedAvg} & \underline{$77.3$} & $49.8$ & $53.73$ \\
  \texttt{FedProx} & $74.9$ & \underline{$50.7$} & \underline{$54.35$} \\
 \texttt{FedNova} & $70.4$ & $46.5$ & $53.61$ \\
 \texttt{SCAFFOLD} & $42.8$ & $49.1$ & $54.15$ \\ \midrule
 \texttt{KAHM} Local Classifier (\ref{eq_170420241923}) & \bfseries 97.95 & \bfseries 97.82 & \bfseries 79.88  \\
 \texttt{LG-FedAvg} & $96.8$ & $86.31$ & $45.98$\\
 \texttt{Per-FedAvg} & $95.95$ & $85.46$ & $60.19$ \\
 \texttt{IFCA} & $97.15$ & $87.99$ & $71.84$ \\
 \texttt{CFL} & $77.93$ & $51.11$ & $40.29$ \\
 \texttt{PACFL} & \underline{$97.54$} & \underline{$89.3$} & \underline{$73.10$} \\
  \bottomrule 
\end{tabular}
\caption{Comparison of the averaged (over clients) test data accuracies obtained by our proposed method with previously available results~\cite{10.1609/aaai.v37i8.26197} for different datasets in the non-iid label skew 20\% federated learning experiment. The mean value of averaged accuracy over 3 independent runs of the experiment is reported.}
\label{tab:2}
\end{table}
\begin{table}
\centering
\begin{tabular}{cccc}
\toprule
 Method & Fashion MNIST & CIFAR-10 & CIFAR-100 \\
 \midrule
\texttt{KAHM} Global Classifier (\ref{eq_150120241845}) & \bfseries 92.75  & \bfseries 95.06  & \bfseries 80.12  \\
  \texttt{FedAvg} & 80.7 & \underline{58.3} & 54.73 \\
  \texttt{FedProx} & \underline{82.5} & 57.1 & 53.31 \\
 \texttt{FedNova} & 78.9 & 54.4 & 54.62 \\
 \texttt{SCAFFOLD} & 77.7 & 57.8 & \underline{54.9} \\ \midrule
 \texttt{KAHM} Local Classifier (\ref{eq_170420241923}) & \bfseries 95.51  & \bfseries 95.90  & \bfseries 73.44  \\
 \texttt{LG-FedAvg} & 94.21 & 76.58 & 35.91 \\
 \texttt{Per-FedAvg} & 92.87 & 77.67 & 56.42 \\
 \texttt{IFCA} & 95.22 & 80.95 & 67.39 \\
 \texttt{CFL} & 78.44 & 52.57 & 35.23 \\
 \texttt{PACFL} & \underline{95.46} & \underline{82.77} & \underline{67.71} \\
  \bottomrule 
\end{tabular}
\caption{Comparison of the averaged (over clients) test data accuracies obtained by our proposed method with the previously available results~\cite{10.1609/aaai.v37i8.26197} for different datasets in the non-iid label skew 30\% federated learning experiment. The mean value of averaged accuracy over 3 independent runs of the experiment is reported.}
\label{tab:3}
\end{table}
The proposed KAHM-based federated learning approach, unlike the state of the art federated learning methods, is outside the realm of gradient descent-based learning of parametric neural networks. Therefore, the performance of the proposed method is evaluated and compared with previously available results~\cite{10.1609/aaai.v37i8.26197} of existing methods. Specifically, the proposed global classifier (\ref{eq_150120241845}) is compared with the methods (that train a single global model across all clients): \texttt{FedAvg}~\shortcite{pmlr-v54-mcmahan17a}, \texttt{FedProx}~\shortcite{DBLP:conf/mlsys/LiSZSTS20}, \texttt{FedNova}~\shortcite{10.5555/3495724.3496362}, and \texttt{SCAFFOLD}~\shortcite{pmlr-v119-karimireddy20a}. The proposed local classifier (\ref{eq_170420241923}) is compared with personalised federated learning methods: \texttt{LG-FedAvg}~\shortcite{DBLP:journals/corr/abs-2001-01523}, \texttt{Per-FedAvg}~\shortcite{NEURIPS2020_24389bfe}, \texttt{CFL}~\shortcite{9174890}, \texttt{IFCA}~\shortcite{10.5555/3495724.3497367}, and \texttt{PACFL}~\cite{10.1609/aaai.v37i8.26197}. Table~\ref{tab:2} reports the results for non-iid label skew 20\% and results for non-iid label skew 30\% are reported in Table~\ref{tab:3}. 

It can be seen from Table~\ref{tab:2} and Table~\ref{tab:3} that our KAHM-based approach consistently achieves considerably better performance, in other words, it visibly outperforms the state of the art federated learning methods. In particular, the KAHM global classifier improved the best existing performance on CIFAR-100 by $+9.3\%$ (in the non-iid label skew 20\% scenario) and by $+12.41\%$ (in the non-iid label skew 30\% scenario). Similarly, the KAHM global classifier improved the best existing performance on CIFAR-10 by $+7.02\%$ (in the non-iid label skew 20\% scenario) and by $+12.29\%$ (in the non-iid label skew 30\% scenario).           
\subsection{Transfer Learning in a Federated Setting}\label{sec_260420241120}
To evaluate the performance of the proposed KAHM-based approach to collaborative learning in a federated setting, we study the performance of global classifier (\ref{eq_150120241845}) in transferring knowledge from a client (corresponding to the source domain) to another client (corresponding to the target domain). For this, we consider the Office-Caltech-10 dataset consisting of four domains: Amazon, Caltech, DSLR, and Webcam. This dataset has been widely used in the literature, e.g.,~\shortcite{Hoffman2013EfficientLO,Herath_2017_CVPR,8362683,Hoffman2014,kumar2023differentially}, for evaluating multi-class accuracy performance in a standard domain adaptation setting with a small number of labelled target samples. We follow the experimental setup of previous studies such as~\shortcite{Hoffman2013EfficientLO,Herath_2017_CVPR,8362683,Hoffman2014,kumar2023differentially} on semi-supervised transfer learning using the Office-Caltech-10 dataset:    
\begin{itemize}
\item the number of training samples per class in the source domain is 20 for Amazon and is 8 for Caltech, DSLR, and Webcam;
\item the number of labelled samples per class in the target domain is 3 for all the four domains;
\item 20 random train/test splits are created and the performance on target domain test samples is averaged over 20 experiments.  
\end{itemize}      
\begin{table}
\centering
 \begin{tabular}{cccc}  
    \toprule
  \bfseries Method & \bfseries Feature Type &  \bfseries Accuracy (\%) & \bfseries Rank  \\  
    \midrule
  \texttt{KAHM} Global Classifier (\ref{eq_150120241845})  & RESNET50 & \textbf{94.3 $\pm$ 3.9} & \bfseries 1 \\   
 \texttt{KAHM} Global Classifier (\ref{eq_150120241845})  & VGG-FC6 & \underline{93.8 $\pm$ 3.9} & \underline{2} \\
\texttt{CDMMA}   & VGG-FC6  & 88.4 $\pm$ 4.3 & 3 \\
\texttt{ILS} (1-NN) & VGG-FC6 &  88.4 $\pm$ 4.4 & 4  \\
\texttt{CDLS} & VGG-FC6 & 85.9 $\pm$ 4.9 &5   \\
\texttt{MMDT} & VGG-FC6 & 80.8 $\pm$ 4.9 & 7 \\
\texttt{HFA} & VGG-FC6 &  83.7 $\pm$ 5.3 & 6 \\
\texttt{OBTL} & SURF &  58.9 $\pm$ 14.6 & 8   \\
\texttt{ILS} (1-NN) & SURF & 55.6 $\pm$ 12.0 & 9   \\
\texttt{CDLS} & SURF &  53.5 $\pm$ 13.0 & 10 \\
\texttt{MMDT} & SURF & 52.5 $\pm$ 13.7 & 11  \\
\texttt{HFA} & SURF & 48.1 $\pm$ 12.0 & 12  \\
        \bottomrule
    \end{tabular}  
 \caption{A summary of the results obtained by different methods in semi-supervised transfer learning experiments on the Office-Caltech-10 dataset. The mean and standard deviation of averaged accuracy over 12 different transfer learning experiments are reported.}
 \label{tab:4}
\end{table}
We consider the following existing method for a comparison: Invariant Latent Space \texttt{ILS} (1-NN)~\shortcite{Herath_2017_CVPR}, Cross-Domain Landmark Selection \texttt{CDLS}~\cite{7780918}, Maximum Margin Domain Transform \texttt{MMDT}~\cite{Hoffman2014}, Heterogeneous Feature Augmentation \texttt{HFA}~\cite{6587717}, Optimal Bayesian Transfer Learning \texttt{OBTL}~\cite{8362683}, and Conditionally Deep Membership-Mapping Autoencoder \texttt{CDMMA}~\cite{kumar2023differentially}. For a fair comparison, the proposed method is also studied with the deep-net VGG-FC6 features extracted from the images, as in the previous studies~\cite{Herath_2017_CVPR,kumar2023differentially}. The Office-Caltech-10 dataset has also been previously studied using SURF features~\cite{Hoffman2013EfficientLO,Herath_2017_CVPR,8362683,Hoffman2014}, and thus existing results using SURF features are additionally considered for a comparison. Taking a domain as source and other domain as target, 12 different transfer learning experiments can be performed on the 4 domains of Office-Caltech-10 dataset. All tables from Table~\ref{table_amazon_2_caltech} - 
Table~\ref{table_webcam_2_dslr} report the results with the two best performances highlighted. The results of all 12 experiments have been summarised in Table~\ref{tab:4}. 
\begin{table}
\centering
  {%
    \begin{tabular}{ccc}  
    \toprule
  \bfseries Method & \bfseries Feature Type &  \bfseries Accuracy (\%)  \\  
    \midrule
  \texttt{KAHM} Global Classifier (\ref{eq_150120241845})  & RESNET50 & \bfseries 89.3 \\   
 \texttt{KAHM} Global Classifier (\ref{eq_150120241845})  & VGG-FC6 & \underline{88.9} \\
\texttt{CDMMA}   & VGG-FC6  & 80.6  \\
\texttt{ILS} (1-NN) & VGG-FC6 &  83.3  \\
\texttt{CDLS} & VGG-FC6 &  78.1  \\
\texttt{MMDT} & VGG-FC6 & 78.7  \\
\texttt{HFA} & VGG-FC6 &  75.5  \\
\texttt{OBTL} & SURF &  41.5  \\
\texttt{ILS} (1-NN) & SURF &  43.6  \\
\texttt{CDLS} & SURF &  35.3  \\
\texttt{MMDT} & SURF &  36.4 \\
\texttt{HFA} & SURF & 31.0  \\
        \bottomrule
    \end{tabular}  
     \caption{Accuracy (averaged over 20 experiments) in Amazon$\rightarrow$Caltech experiments.}
 \label{table_amazon_2_caltech}
  }  
\end{table}  
\begin{table}
\centering
  {%
    \begin{tabular}{ccc}  
    \toprule 
  \bfseries Method & \bfseries Feature Type &  \bfseries Accuracy (\%)  \\  
    \midrule
      \texttt{KAHM} Global Classifier (\ref{eq_150120241845})  & RESNET50 & \bfseries 95.7 \\   
 \texttt{KAHM} Global Classifier (\ref{eq_150120241845})  & VGG-FC6 & \underline{95.6} \\
\texttt{CDMMA}   & VGG-FC6 &   91.2      \\
\texttt{ILS} (1-NN) & VGG-FC6 &  87.7 \\
\texttt{CDLS} & VGG-FC6 &  86.9   \\
\texttt{MMDT} & VGG-FC6 &  77.1   \\
\texttt{HFA} & VGG-FC6 &  87.1   \\
\texttt{OBTL} & SURF &  60.2   \\
\texttt{ILS} (1-NN) & SURF &  49.8   \\
\texttt{CDLS} & SURF &  60.4   \\
\texttt{MMDT} & SURF &  56.7   \\
\texttt{HFA} & SURF &  55.1  \\
        \bottomrule
    \end{tabular}  
     \caption{Accuracy (averaged over 20 experiments) in Amazon$\rightarrow$DSLR experiments.}
 \label{table_amazon_2_dslr}
  }  
\end{table} 
\begin{table}
\centering
  {%
    \begin{tabular}{ccc}  
    \toprule
  \bfseries Method & \bfseries Feature Type & \bfseries Accuracy (\%)  \\  
    \midrule
          \texttt{KAHM} Global Classifier (\ref{eq_150120241845})  & RESNET50 & \bfseries 96.7\\   
 \texttt{KAHM} Global Classifier (\ref{eq_150120241845})  & VGG-FC6 & \underline{95.5} \\
\texttt{CDMMA}  & VGG-FC6 &     89.5     \\
\texttt{ILS} (1-NN) & VGG-FC6 &  90.7   \\
\texttt{CDLS} & VGG-FC6 &  91.2  \\
\texttt{MMDT} & VGG-FC6 &  82.5   \\
\texttt{HFA} & VGG-FC6 &  87.9   \\
\texttt{OBTL} & SURF &  72.4   \\
\texttt{ILS} (1-NN) & SURF &  59.7   \\
\texttt{CDLS} & SURF &  68.7   \\
\texttt{MMDT} & SURF &   64.6  \\
\texttt{HFA} & SURF &  57.4  \\
        \bottomrule
    \end{tabular}  
    \caption{Accuracy (averaged over 20 experiments) in Amazon$\rightarrow$Webcam experiments.}
 \label{table_amazon_2_webcam}
  }  
\end{table}   
\begin{table}
\centering
  {%
    \begin{tabular}{ccc}  
    \toprule 
  \bfseries Method & \bfseries Feature Type &  \bfseries Accuracy (\%)  \\  
    \midrule
      \texttt{KAHM} Global Classifier (\ref{eq_150120241845})  & RESNET50 & \bfseries 93.9 \\   
 \texttt{KAHM} Global Classifier (\ref{eq_150120241845})  & VGG-FC6 & \underline{93.6} \\
\texttt{CDMMA}  & VGG-FC6 &    91.5   \\
\texttt{ILS} (1-NN) & VGG-FC6 &  89.7  \\
\texttt{CDLS} & VGG-FC6 &  88.0  \\
\texttt{MMDT} & VGG-FC6 &  85.9  \\
\texttt{HFA} & VGG-FC6 &  86.2   \\
\texttt{OBTL} & SURF &   54.8  \\
\texttt{ILS} (1-NN) & SURF &   55.1   \\
\texttt{CDLS} & SURF &   50.9  \\
\texttt{MMDT} & SURF &  49.4   \\
\texttt{HFA} & SURF &  43.8  \\
        \bottomrule
    \end{tabular}  
 \caption{Accuracy (averaged over 20 experiments) in Caltech$\rightarrow$Amazon experiments.}
 \label{table_caltech_2_amazon}
  }  
\end{table}   
\begin{table}
\centering
  {%
    \begin{tabular}{ccc}  
    \toprule
  \bfseries Method & \bfseries Feature Type &  \bfseries Accuracy (\%)  \\  
  \midrule
      \texttt{KAHM} Global Classifier (\ref{eq_150120241845})  & RESNET50 & \bfseries 96.5 \\   
 \texttt{KAHM} Global Classifier (\ref{eq_150120241845})  & VGG-FC6 & \underline{95.4} \\
 \texttt{CDMMA}  & VGG-FC6 &    91.6 \\
\texttt{ILS} (1-NN) & VGG-FC6 &   86.9  \\
\texttt{CDLS} & VGG-FC6 &  86.3  \\
\texttt{ MMDT} & VGG-FC6 &   77.9   \\
\texttt{ HFA} & VGG-FC6 &  87.0    \\
\texttt{ OBTL} & SURF &  61.5   \\
\texttt{ILS} (1-NN) & SURF &   56.2  \\
\texttt{CDLS} & SURF &   59.8  \\
\texttt{MMDT} & SURF &   56.5  \\
\texttt{HFA} & SURF &   55.6 \\
        \bottomrule
    \end{tabular}  
  \caption{Accuracy (averaged over 20 experiments) in Caltech$\rightarrow$DSLR experiments.}
 \label{table_caltech_2_dslr}
  }  
\end{table} 
\begin{table}
\centering
  {%
    \begin{tabular}{ccc}  
    \toprule
  \bfseries Method & \bfseries Feature Type &  \bfseries Accuracy (\%)  \\  
 \midrule
      \texttt{KAHM} Global Classifier (\ref{eq_150120241845})  & RESNET50 & \bfseries 96.0 \\   
 \texttt{KAHM} Global Classifier (\ref{eq_150120241845})  & VGG-FC6 & \underline{95.0} \\
 \texttt{CDMMA}    & VGG-FC6 &    91.6   \\
 \texttt{ ILS} (1-NN) & VGG-FC6 &  91.4   \\
 \texttt{ CDLS} & VGG-FC6 &  89.7   \\
 \texttt{ MMDT} & VGG-FC6 &  82.8   \\
 \texttt{HFA }& VGG-FC6 &  86.0    \\
 \texttt{ OBTL} & SURF &  71.1   \\
 \texttt{ ILS} (1-NN) & SURF &   62.9  \\
 \texttt{ CDLS} & SURF &  66.3   \\
 \texttt{MMDT} & SURF &   63.8  \\
 \texttt{ HFA} & SURF &  58.1  \\
        \bottomrule 
    \end{tabular}  
  \caption{Accuracy (averaged over 20 experiments) in Caltech$\rightarrow$Webcam experiments.}
 \label{table_caltech_2_webcam}
  }  
\end{table}   
 \begin{table}
\centering
  {%
    \begin{tabular}{ccc}  
    \toprule 
  \bfseries Method & \bfseries Feature Type &  \bfseries Accuracy (\%)   \\  
   \midrule
      \texttt{KAHM} Global Classifier (\ref{eq_150120241845})  & RESNET50 & \bfseries 94.0 \\   
 \texttt{KAHM} Global Classifier (\ref{eq_150120241845})  & VGG-FC6 & \underline{93.7} \\
 \texttt{CDMMA}     & VGG-FC6 &   90.7 \\
 \texttt{ILS} (1-NN) & VGG-FC6 & 88.7  \\
 \texttt{CDLS} & VGG-FC6 &   88.1  \\
 \texttt{MMDT} & VGG-FC6 &   83.6  \\
 \texttt{HFA} & VGG-FC6 &   85.9   \\
 \texttt{OBTL} & SURF &  54.4  \\
 \texttt{ILS} (1-NN) & SURF &   55.0  \\
 \texttt{CDLS} & SURF &  50.7  \\
 \texttt{ MMDT} & SURF &  46.9  \\
 \texttt{HFA} & SURF &   42.9  \\
        \bottomrule
    \end{tabular}  
  \caption{Accuracy (averaged over 20 experiments) in DSLR$\rightarrow$Amazon experiments.}
 \label{table_dslr_2_amazon}
  }  
\end{table}    
\begin{table}
\centering
  {%
    \begin{tabular}{ccc}  
    \toprule 
  \bfseries Method & \bfseries Feature Type & \bfseries Accuracy (\%)  \\  
 \midrule
      \texttt{KAHM} Global Classifier (\ref{eq_150120241845})  & RESNET50 & \bfseries 88.5 \\   
 \texttt{KAHM} Global Classifier (\ref{eq_150120241845})  & VGG-FC6 & \underline{87.9} \\
 \texttt{CDMMA}    & VGG-FC6 &    81.4   \\
 \texttt{ ILS} (1-NN) & VGG-FC6 &   81.4  \\
 \texttt{CDLS} & VGG-FC6 &  77.9   \\
 \texttt{ MMDT} & VGG-FC6 &  71.8  \\
 \texttt{HFA} & VGG-FC6 &  74.8   \\
 \texttt{OBTL} & SURF &  40.3  \\
 \texttt{ ILS} (1-NN) & SURF &  41.0  \\
 \texttt{CDLS} & SURF &  34.9  \\
 \texttt{MMDT} & SURF &  34.1  \\
 \texttt{HFA} & SURF &   30.9  \\
        \bottomrule
    \end{tabular}  
  \caption{Accuracy (averaged over 20 experiments) in DSLR$\rightarrow$Caltech experiments.}
 \label{table_dslr_2_caltech}
  }  
\end{table}   
 \begin{table}
\centering
  {%
    \begin{tabular}{cccc}  
  \toprule 
  \bfseries Method & \bfseries Feature type & \bfseries Accuracy (\%)  \\  
 \midrule
      \texttt{KAHM} Global Classifier (\ref{eq_150120241845})  & RESNET50 & \bfseries 99.4 \\   
 \texttt{KAHM} Global Classifier (\ref{eq_150120241845})  & VGG-FC6 & \underline{98.8} \\
 \texttt{CDMMA} & VGG-FC6 &   88.7    \\
 \texttt{ ILS} (1-NN) & VGG-FC6 &   95.5  \\
 \texttt{ CDLS } & VGG-FC6 &  90.7  \\
 \texttt{ MMDT} & VGG-FC6 &   86.1 \\
 \texttt{HFA} & VGG-FC6 &   86.9  \\
 \texttt{ OBTL} & SURF &  83.2 \\
 \texttt{ ILS} (1-NN) & SURF &  80.1   \\
 \texttt{ CDLS} & SURF &  68.5  \\
 \texttt{ MMDT} & SURF &  74.1  \\
 \texttt{ HFA }& SURF &  60.5   \\
        \bottomrule
    \end{tabular}  
  \caption{Accuracy (averaged over 20 experiments) in DSLR$\rightarrow$Webcam experiments.}
 \label{table_dslr_2_webcam}
  }  
\end{table}   
\begin{table}
\centering
  {%
    \begin{tabular}{cccc}  
  \toprule 
  \bfseries Method & \bfseries Feature Type & \bfseries Accuracy (\%)  \\  
 \midrule
      \texttt{KAHM} Global Classifier (\ref{eq_150120241845})  & RESNET50 & \bfseries 94.3 \\   
 \texttt{KAHM} Global Classifier (\ref{eq_150120241845})  & VGG-FC6 & \underline{94.0} \\
 \texttt{CDMMA} & VGG-FC6 &   92.0    \\
 \texttt{ ILS} (1-NN) & VGG-FC6 &  88.8   \\
 \texttt{ CDLS} & VGG-FC6 &  87.4  \\
 \texttt{ MMDT} & VGG-FC6 &  84.7  \\
 \texttt{ HFA} & VGG-FC6 &  85.1  \\
 \texttt{ OBTL} & SURF &   55.0 \\
 \texttt{ ILS} (1-NN) & SURF & 54.3   \\
 \texttt{ CDLS} & SURF &  51.8  \\
 \texttt{ MMDT} & SURF &  47.7  \\
 \texttt{ HFA} & SURF &  56.5   \\
        \bottomrule
    \end{tabular}  
  \caption{Accuracy (averaged over 20 experiments) in Webcam$\rightarrow$Amazon experiments.}
 \label{table_webcam_2_amazon}
  }  
\end{table}   
 \begin{table}
\centering
  {%
    \begin{tabular}{cccc}  
  \toprule 
  \bfseries Method & \bfseries Feature Type & \bfseries Accuracy (\%)  \\  
 \midrule
      \texttt{KAHM} Global Classifier (\ref{eq_150120241845})  & RESNET50 & \bfseries 88.2 \\   
 \texttt{KAHM} Global Classifier (\ref{eq_150120241845})  & VGG-FC6 & \underline{87.3} \\
 \texttt{CDMMA}    & VGG-FC6 &   82.3    \\
 \texttt{ILS} (1-NN) & VGG-FC6 &  82.8   \\
 \texttt{CDLS} & VGG-FC6 &  78.2  \\
 \texttt{MMDT} & VGG-FC6 &  73.6  \\
 \texttt{HFA} & VGG-FC6 &  74.4   \\
 \texttt{OBTL} & SURF &  37.4   \\
 \texttt{ ILS} (1-NN) & SURF &   38.6   \\
 \texttt{ CDLS} & SURF &  33.5   \\
 \texttt{MMDT} & SURF & 32.2   \\
 \texttt{HFA} & SURF &   29.0  \\
        \bottomrule
    \end{tabular}  
   \caption{Accuracy (averaged over 20 experiments) in Webcam$\rightarrow$Caltech experiments.}
 \label{table_webcam_2_caltech}
  }  
\end{table}
\begin{table}
\centering
  {%
    \begin{tabular}{ccc}  
    \toprule 
  \bfseries Method & \bfseries Feature Type & \bfseries Accuracy (\%)  \\  
 \midrule
      \texttt{KAHM} Global Classifier (\ref{eq_150120241845})  & RESNET50 & \bfseries 99.6 \\   
 \texttt{KAHM} Global Classifier (\ref{eq_150120241845})  & VGG-FC6 & \bfseries 99.6 \\
 \texttt{CDMMA}    & VGG-FC6 &    89.6     \\
 \texttt{ ILS} (1-NN) & VGG-FC6 &  \underline{94.5}  \\
 \texttt{ CDLS} & VGG-FC6 &  88.5   \\
 \texttt{ MMDT} & VGG-FC6 &   85.1  \\
 \texttt{ HFA} & VGG-FC6 &   87.3 \\
 \texttt{ OBTL} & SURF &  75.0   \\
 \texttt{ ILS} (1-NN) & SURF  & 70.8    \\
 \texttt{ CDLS} & SURF &  60.7  \\
 \texttt{ MMDT} & SURF &   67.0  \\
 \texttt{ HFA} & SURF &   56.5  \\
        \bottomrule
    \end{tabular}  
  \caption{Accuracy (averaged over 20 experiments) in Webcam$\rightarrow$DSLR experiments.}
 \label{table_webcam_2_dslr}
  }  
\end{table}

The best performance of our proposed method is observed in Table~\ref{tab:4} and consistently in all of the 12 experiments (as observed from Table~\ref{table_amazon_2_caltech} 
through to
Table~\ref{table_webcam_2_dslr}). Overall, as observed from Table~\ref{tab:4}, the KAHM global classifier improved the best existing performance on Office-Caltech-10 dataset by $+5.9\%$ using ResNet-50 features and by $+5.4\%$ using VGG16 features. 
\subsection{Single-Class Data}\label{sec_120520242021}
The Freiburg Groceries dataset is again considered under the scenario that a client owns training samples of only one class and all of the samples of that class, resulting in $Q = C$. As in the previous studies on this dataset \shortcite{10.1007/978-3-030-87101-7_14,8888203,KUMAR20211}, the experiments are performed for 5 different train-test splits of data. The obtained results are reported in Table~\ref{table_results_grocery_images} and compared with the previous results on this dataset. The improved performance shown in Table~\ref{table_results_grocery_images} proves that the proposed method remains competitive under the considered single-class data scenario.  \begin{table}[h]
\centering
\begin{tabular}{c|*{6}{c}}
\toprule
\bfseries \multirow{2}{*}{\bfseries methods} & \multicolumn{6}{c}{\bfseries accuracy (in \%) on test images}  \\
\cline{2-7} 
 &   1 &   2 &   3 &   4 &   5 &  mean \\ 
\hline \hline 
 \texttt{KAHM} Global Classifier (\ref{eq_150120241845}) & \bfseries 89.69 & \bfseries 87.79 & \bfseries 88.02 & \bfseries 88.44 & \bfseries 87.99 & \bfseries 88.39 \\ \hline 
$\begin{array}{c} \mbox{membership-mappings} \\ \mbox{\cite{10.1007/978-3-030-87101-7_14}} \end{array}$ & 87.82 & \underline{87.06}  & \underline{85.88}  & \underline{85.63} & \underline{86.19} & \underline{86.52}  \\ \hline
$\begin{array}{c} \mbox{nonparametric fuzzy} \\ \mbox{image mapping}\\ \mbox{\cite{8888203}} \end{array}$ & \underline{88.21} & 86.64  & 85.36  & 85.13 & 85.79 &   86.23 \\ \hline
$\begin{array}{c} \mbox{Gaussian fuzzy-mapping} \\ \mbox{\cite{KUMAR20211}} \end{array}$ & 83.50  & 81.52  & 79.73  & 79.60  & 80.48  & 80.97  \\ \hline
SVM~\cite{10.1007/978-3-030-87101-7_14} & 77.90   & 79.54    & 77.17    & 76.98     & 76.98 & 77.71  \\ \hline
$1$-NN~\cite{10.1007/978-3-030-87101-7_14} & 78.00   & 77.97    & 77.38    & 76.58    & 76.28 & 77.24   \\ \hline
$\begin{array}{c} \mbox{Back-propagation} \\ \mbox{training of a deep network} \\ \mbox{\cite{10.1007/978-3-030-87101-7_14}} \end{array}$ & 75.25 & 77.24 & 72.67  & 73.37 & 71.57 & 74.02 \\ \hline
$2$-NN~\cite{10.1007/978-3-030-87101-7_14} & 73.48  & 73.38  & 70.11  & 70.05  & 70.57 & 71.52  \\ \hline
$4$-NN~\cite{10.1007/978-3-030-87101-7_14} & 72.50 & 73.39  & 68.89  & 71.16  & 70.87 & 71.36  \\ \hline
$\begin{array}{c} \mbox{Random Forest}\\ \mbox{\cite{10.1007/978-3-030-87101-7_14}} \end{array}$ &  63.17 & 62.63 & 59.47 & 59.50 &  59.76 & 60.90\\ \hline
$\begin{array}{c} \mbox{Naive Bayes} \\ \mbox{\cite{10.1007/978-3-030-87101-7_14}} \end{array}$ & 56.78  & 56.78      & 53.74   & 55.08     & 56.26 & 55.73  \\ \hline
$\begin{array}{c} \mbox{Ensemble Learning} \\ \mbox{\cite{10.1007/978-3-030-87101-7_14}} \end{array}$ & 38.31  & 39.35 & 38.89  &  37.69    & 38.34 & 38.51  \\ \hline
$\begin{array}{c} \mbox{Decision Tree} \\ \mbox{\cite{10.1007/978-3-030-87101-7_14}} \end{array}$ & 31.34   & 30.59    & 32.14    & 31.06    & 30.73 & 31.17  \\ 
\bottomrule
\end{tabular}
\caption{Results of single-class data federated learning experiments on 5 different train-test splits of Freiburg groceries data}
\label{table_results_grocery_images} 
\end{table}
\section{Conclusion}\label{sec_conclusion}
Collaborative learning in federated setting is becoming popular due to the increasingly distributed nature of data across a variety of domains. Overall, we provide a KAHM-based comprehensive solution. Unlike most of the existing methods, our approach does not require the multiple rounds of communication between clients and server for the learning of the global model, does not require clients to perform multiple epochs of local optimisation using stochastic gradient descent, and does not require any tuning of the free parameters. This paper provides a new theory for KAHM-based collaborative learning in a federated setting. Our work sheds light on the theoretical understanding of KAHM-based collaborative learning in federated settings and provides insights for designing a suitable learning solution. We have provided theoretical guarantees for the proposed solution and empirically demonstrated its effectiveness by showing a considerable improvement from the existing results on benchmark datasets in federated learning and transfer learning experiments. An interpretation and justification for the proposed solution is also provided in terms of a distance measure from training data points.       

The KAHM approach paves the way for AutoML. However, this requires addressing a limitation of the current work that the scope of the study does not include representation learning, and thus our future work will study representation learning by expanding a KAHM-based approach to automatically extract features from raw data (including images) at varying abstraction levels. In particular, we will focus on developing KAHM-based representations learning algorithms outside the realm of stochastic gradient descent. The other limitation of the current work is providing theoretical analysis only for the squared loss function, and thus we will also extend the results to a family of loss functions in a generalised setting. Since Rademacher complexity is a traditional technique, it is also worth investigating operator and spectral methods for geometrically inspired kernel machines. Specifically, we will study the effect of spectral properties of the kernel matrix for achieving even tighter bounds on the errors. The current study achieves robustness towards statistical heterogeneity (among clients' data) by design. Another possible extension is to establish and study the robustness of the method towards noise present in the data and quantifying the uncertainties affecting the prediction. Finally, an extension of the methodology to include time series data will be considered in the future.       

\acks{The research reported in this paper has been supported by the state of Upper Austria as part of \#upperVISION2030 under the Secure Prescriptive Analytics (SPA) project; Austrian Ministry for Transport, Innovation and Technology, the Federal Ministry for Digital and Economic Affairs, and the State of Upper Austria in the frame of the SCCH competence center INTEGRATE [(FFG grant no. 892418)] part of the FFG COMET Competence Centers for Excellent Technologies Programme. Bowles is partially  supported by the Austrian Funding Council (FWF) under Meitner M 3338-N.
}

\appendix

\section*{Appendix A. Details of KAHM Definition~\ref{def_affine_hull_model}}
Given a finite number of samples: $\mathbf{Y} = \left[\begin{IEEEeqnarraybox*}[][c]{,c/c/c,} y^1 & \cdots & y^N \end{IEEEeqnarraybox*} \right]^T$ with $y^1,\cdots,y^N \in \mathbb{R}^p$ and a subspace dimension $n \leq p$; a kernel affine hull machine $\mathcal{A}_{\mathbf{Y},n}: \mathbb{R}^p \rightarrow \mathrm{aff}(\{y^1,\cdots,y^N \})$ is defined as
  \begin{IEEEeqnarray}{rCl}
\label{eq_301220221131}\mathcal{A}_{\mathbf{Y},n}(y) & := & \frac{h_{k_{\theta},\mathbf{Y}\mathbf{P}^T,\lambda^*}^1(\mathbf{P}y)}{\sum_{i=1}^Nh_{k_{\theta},\mathbf{Y}\mathbf{P}^T,\lambda^*}^i(\mathbf{P}y)}y^1 + \cdots + \frac{h_{k_{\theta},\mathbf{Y}\mathbf{P}^T,\lambda^*}^N(\mathbf{P}y)}{\sum_{i=1}^Nh_{k_{\theta},\mathbf{Y}\mathbf{P}^T,\lambda^*}^i(\mathbf{P}y)}y^N.
   \end{IEEEeqnarray} 
 Here,
\begin{itemize}
\item $\mathbf{P} \in \mathbb{R}^{n \times p}\: (n \leq p)$ is an encoding matrix such that product $\mathbf{P}y$ is a lower-dimensional encoding for $y$. For a given subspace dimension $n$, $\mathbf{P}$ is defined by setting the $i-$th row of $\mathbf{P}$ as equal to transpose of eigenvector corresponding to $i-$th largest eigenvalue of sample covariance matrix of dataset $\{y^1,\cdots,y^N \}$.     
\item $k_{\theta}: \mathcal{X} \times \mathcal{X} \rightarrow \mathbb{R}$ is a positive-definite real-valued kernel on $\mathcal{X}$ with a corresponding reproducing kernel Hilbert space $\mathcal{H}_{k_{\theta}}(\mathcal{X})$ where 
   \begin{IEEEeqnarray}{rCl} 
\mathcal{X} & = & \{ \mathbf{P}y \; \mid \; y \in \mathbb{R}^p \}.
   \end{IEEEeqnarray}   
 The kernel function $k_{\theta}$ is chosen of Gaussian type:
   \begin{IEEEeqnarray}{rCl}
\label{eq_260120231329}k_{\theta}(x^i,x^j) & := & \exp\left(-\frac{1}{2n}(x^i-x^j)^T\theta^{-1}(x^i-x^j)\right), 
  \end{IEEEeqnarray} 
where $\theta \succ 0$ is sample covariance matrix of dataset $\{\mathbf{P}y^1,\cdots,\mathbf{P}y^N \}$.
\item The function $h^i_{k_{\theta},\mathbf{Y}\mathbf{P}^T,\lambda}: \mathcal{X} \rightarrow \mathbb{R}$, such that $h^i_{k_{\theta},\mathbf{Y}\mathbf{P}^T,\lambda} \in \mathcal{H}_{k_{\theta}}(\mathcal{X})$, approximates the indicator function $\mathbbm{1}_{\{\mathbf{P}y^i\}}: \mathcal{X} \rightarrow \{0,1 \}$ as the solution of following kernel regularized least squares problem:  
  \begin{IEEEeqnarray}{rCl}
h^i_{k_{\theta},\mathbf{Y}\mathbf{P}^T,\lambda} & = & \arg \; \min_{f \in \mathcal{H}_{k_{\theta}}(\mathcal{X})} \; \left( \sum_{j=1}^N \left |\mathbbm{1}_{\{\mathbf{P}y^i\}}(\mathbf{P}y^j) - f(\mathbf{P}y^j) \right |^2 + \lambda \left \| f \right \|^2_{\mathcal{H}_k(\mathcal{X})} \right),\; \lambda \in \mathbb{R}_+. \IEEEeqnarraynumspace
  \end{IEEEeqnarray}
The solution follows as
  \begin{IEEEeqnarray}{rCl}
\label{eq_250220231734} h^i_{k_{\theta},\mathbf{Y}\mathbf{P}^T,\lambda}(\cdot) & = & (\mathbf{I}_N)_{i,:} \left(K_{\mathbf{Y}\mathbf{P}^T} + \lambda \mathbf{I}_N \right)^{-1}  \left[\begin{IEEEeqnarraybox*}[][c]{,c/c/c,} k_{\theta}(\cdot,\mathbf{P}y^1) & \cdots & k_{\theta}(\cdot,\mathbf{P}y^N)\end{IEEEeqnarraybox*} \right]^T
  \end{IEEEeqnarray} 
where $(\mathbf{I}_N)_{i,:}$ denotes the $i-$th row of identity matrix of size $N$ and $\mathbf{K}_{\mathbf{Y}\mathbf{P}^T}$ is $N \times N$ kernel matrix with its $(i,j)-$th element defined as
   \begin{IEEEeqnarray}{rCl}
\label{eq_010120231042}(\mathbf{K}_{\mathbf{Y}\mathbf{P}^T})_{ij}& := & k_{\theta}(\mathbf{P}y^i,\mathbf{P}y^j).
  \end{IEEEeqnarray}
The value $ h^i_{k_{\theta},\mathbf{Y}\mathbf{P}^T,\lambda}(\mathbf{P}y)$ represents the kernel-smoothed membership of point $Py$ to the set $\{ \mathbf{P}y^i\}$. 
\item The regularization parameter $\lambda^* \in \mathbb{R}_+$ is given as
  \begin{IEEEeqnarray}{rCl}
\label{eq_010120231034}\lambda^* & = &  \hat{e} + \frac{2}{pN}\|\mathbf{Y} \|^2_F, 
     \end{IEEEeqnarray}   
where $\hat{e}$ is the unique fixed point of $\mathcal{R}_{k_{\theta},\mathbf{Y}\mathbf{P}^T,\mathbf{Y}}$ such that
  \begin{IEEEeqnarray}{rCl}
\label{eq_090120230831}\hat{e} & = & \mathcal{R}_{k_{\theta},\mathbf{Y}\mathbf{P}^T,\mathbf{Y}}(\hat{e},\frac{2}{pN}\|\mathbf{Y} \|^2_F),
     \end{IEEEeqnarray}   
with $\mathcal{R}_{k_{\theta},\mathbf{Y}\mathbf{P}^T,\mathbf{Y}}: \mathbb{R}_{+} \times \mathbb{R}_{+} \rightarrow \mathbb{R}_{+}$ defined as
   \begin{IEEEeqnarray}{rCl}
\label{eq_090120230832}\mathcal{R}_{k_{\theta},\mathbf{Y}\mathbf{P}^T,\mathbf{Y}}(e,\tau)& := &   \frac{1}{pN} \sum_{j=1}^p \|(\mathbf{Y})_{:,j} - \mathbf{K}_{\mathbf{Y}\mathbf{P}^T} \left(\mathbf{K}_{\mathbf{Y}\mathbf{P}^T} + (e+\tau) \mathbf{I}_N \right)^{-1} (\mathbf{Y})_{:,j}\|^2.
     \end{IEEEeqnarray} 
The following iterations
 \begin{IEEEeqnarray}{rCl}
e|_{it+1} & = & \mathcal{R}_{k_{\theta},\mathbf{Y}\mathbf{P}^T,\mathbf{Y}}(e|_{it},\frac{2}{pN}\|\mathbf{Y} \|^2_F),\; it \in \{0,1,\cdots \}  \\
e|_0 & \in & (0,\frac{1}{pN} \|\mathbf{Y} \|^2_F)  
  \end{IEEEeqnarray}
converge to $\hat{e}$.     
\item The image of $\mathcal{A}_{\mathbf{Y},n}$ defines a region in the affine hull of $\{y^1,\cdots,y^N\}$. That is,
 \begin{IEEEeqnarray}{rCCCl}
\label{eq_738825.6428} \mathcal{A}_{\mathbf{Y},n}[\mathbb{R}^p]& := & \{ \mathcal{A}_{\mathbf{Y},n}(y) \; \mid \; y \in \mathbb{R}^p  \}  & \subset & \mathrm{aff}(\{y^1,\cdots,y^N \}).  
  \end{IEEEeqnarray}   
\end{itemize}
\section*{Appendix B. Practical Choice for KAHM Subspace Dimension}
Given $N$ number of samples, the subspace dimension $n$ can not exceed data dimension $p$ and $N-1$ (as the number of principal components with non-zero variance cannot exceed $N-1$). Further, $n$ should not be too high to cause negligible variance along any of the principal components. This can be ensured by checking data variance along each principal component, and if needed decrementing $n$ by 1 till required. Following algorithm is suggested to practically determine $n$: 
\begin{algorithmic}[1]
\Require Dataset $\{y^i \in \mathbb{R}^p\}_{i=1}^N$.
\State $n \gets \min(20,p,N-1)$.
\State  Define $\mathbf{P} \in \mathbb{R}^{n \times p}$ such that the $i-$th row of $\mathbf{P}$ is equal to transpose of eigenvector corresponding to $i-$th largest eigenvalue of sample covariance matrix of samples $\{y^1,\cdots,y^N \}$.
\State Define $x^i = \mathbf{P}y^i,\; \forall i \in \{1,2,\cdots,N\}$. 
\While{$\mathop{\min}_{1\leq j \leq p}\left( \mathop{\max}_{1 \leq i \leq N} (x^i)_j  - \mathop{\min}_{1 \leq i \leq N} (x^i)_j \right) < 1\mathrm{e}{-3}$}
\State  $n \gets n-1$.
\State  Define $\mathbf{P} \in \mathbb{R}^{n \times p}$ such that the $i-$th row of $\mathbf{P}$ is equal to transpose of eigenvector corresponding to $i-$th largest eigenvalue of sample covariance matrix of dataset $\{y^1,\cdots,y^N \}$.
\State Define $x^i = \mathbf{P}y^i,\; \forall i \in \{1,2,\cdots,N\}$. 
\EndWhile
\State \Return $n$
\end{algorithmic}   
\section*{Appendix C. Proof of Theorem~\ref{result_rademacher_complexity_hypthesis_space}}
 Consider
 \begin{IEEEeqnarray}{rCl}
\widehat{\mathcal{R}}_{\mathcal{D}}(\mathcal{M}_{\mathcal{D},c}) & = & \frac{1}{N} \mathop{\mathbb{E}}_{\sigma } \left[ \sup_{f_{y \mapsto z_c} \in \mathcal{M}_{\mathcal{D},c} } \sum_{i=1}^N \sigma_i \: f_{y \mapsto z_c}(y^i) \right] \\
\label{eq_090420241541}& = & \frac{1}{N} \mathop{\mathbb{E}}_{\sigma } \left[ \sup_{f_{y \mapsto z_c} \in \mathcal{M}_{\mathcal{D},c} } \sum_{i=1}^N \sigma_i \: \left \langle f_{y \mapsto z_c}, \mathcal{K}_{c}(\cdot,y^i) \right \rangle_{ \mathcal{H}_{\mathcal{K}_{c}}(\mathbb{R}^p)} \right] \\
& = & \frac{1}{N} \mathop{\mathbb{E}}_{\sigma } \left[ \sup_{f_{y \mapsto z_c} \in \mathcal{M}_{\mathcal{D},c} }  \left \langle f_{y \mapsto z_c}, \sum_{i=1}^N \sigma_i \mathcal{K}_{c}(\cdot,y^i) \right \rangle_{ \mathcal{H}_{\mathcal{K}_{c}}(\mathbb{R}^p) } \right],
  \end{IEEEeqnarray} 
where (\ref{eq_090420241541}) follows from the reproducing property of the kernel $\mathcal{K}_{c}$. Using Cauchy–Schwarz inequality,
\begin{IEEEeqnarray}{rCl}
 \widehat{\mathcal{R}}_{\mathcal{D}}(\mathcal{M}_{\mathcal{D},c}) & \leq  & \frac{1}{N} \mathop{\mathbb{E}}_{\sigma } \left[ \sup_{f_{y \mapsto z_c} \in \mathcal{M}_{\mathcal{D},c}}    \|f_{y \mapsto z_c}  \|_{ \mathcal{H}_{\mathcal{K}_{c}}(\mathbb{R}^p)} \left \| \sum_{i=1}^N \sigma_i \mathcal{K}_{c}(\cdot,y^i) \right \|_{ \mathcal{H}_{\mathcal{K}_{c}}(\mathbb{R}^p)}   \right], 
 \end{IEEEeqnarray} 
 and using (\ref{eq_150120241129}), we have
\begin{IEEEeqnarray}{rCl}
 \widehat{\mathcal{R}}_{\mathcal{D}}(\mathcal{M}_{\mathcal{D},c}) & \leq &  \frac{1}{N} \mathop{\mathbb{E}}_{\sigma } \left[  \left \| \sum_{i=1}^N \sigma_i \mathcal{K}_{c}(\cdot,y^i) \right \|_{\mathcal{H}_{\mathcal{K}_{c}}(\mathbb{R}^p)}   \right].
 \end{IEEEeqnarray} 
 As per Jensen's inequality
 \begin{IEEEeqnarray}{rCl}
 \left( \mathop{\mathbb{E}}_{\sigma } \left[  \left \| \sum_{i=1}^N \sigma_i \mathcal{K}_{c}(\cdot,y^i) \right \|_{\mathcal{H}_{\mathcal{K}_{c}}(\mathbb{R}^p)}   \right] \right)^2 & \leq & \mathop{\mathbb{E}}_{\sigma } \left[  \left \| \sum_{i=1}^N \sigma_i \mathcal{K}_{c}(\cdot,y^i) \right \|_{\mathcal{H}_{\mathcal{K}_{c}}(\mathbb{R}^p)}^2   \right],
 \end{IEEEeqnarray} 
 and thus
\begin{IEEEeqnarray}{rCl}
\widehat{\mathcal{R}}_{\mathcal{D}}(\mathcal{M}_{\mathcal{D},c})  & \leq  &  \frac{1}{N}  \sqrt{\mathop{\mathbb{E}}_{\sigma } \left[  \left \| \sum_{i=1}^N \sigma_i \mathcal{K}_{c}(\cdot,y^i) \right \|_{ \mathcal{H}_{\mathcal{K}_{c}}(\mathbb{R}^p)}^2   \right]} \\
& = &   \frac{1}{N}  \sqrt{\mathop{\mathbb{E}}_{\sigma } \left[ \left \langle \sum_{i=1}^N \sigma_i \mathcal{K}_{c}(\cdot,y^i), \sum_{i=1}^N \sigma_i \mathcal{K}_{c}(\cdot,y^i) \right \rangle _{ \mathcal{H}_{\mathcal{K}_{c}}(\mathbb{R}^p)}   \right]} \\
& = &  \frac{1}{N}   \sqrt{\mathop{\mathbb{E}}_{\sigma } \left[ \sum_{i,j=1}^N \sigma_i \sigma_j  \mathcal{K}_{c}(y^i,y^j)    \right]} \\
& = &  \frac{1}{N}  \sqrt{ \sum_{i,j=1}^N  \mathcal{K}_{c}(y^i,y^j) \mathop{\mathbb{E}}_{\sigma } \left[ \sigma_i \sigma_j \right]     }. 
 \end{IEEEeqnarray} 
 Since $\sigma_1,\cdots,\sigma_N$ are independent random variables drawn from the Rademacher distribution, we have
\begin{IEEEeqnarray}{rCCCl}
 \widehat{\mathcal{R}}_{\mathcal{D}}(\mathcal{M}_{\mathcal{D},c}) & \leq  &  \frac{1}{N} \sqrt{ \sum_{i=1}^N  \mathcal{K}_{c}(y^i,y^i)      } & \leq &  \frac{1}{\sqrt{N}}.
\end{IEEEeqnarray} 
Hence, (\ref{eq_150120241133}) and (\ref{eq_150120241205}) follow.
\section*{Appendix D. Proof of Lemma~\ref{result_rademacher_complexity_loss_space}}
The empirical Rademacher complexity of $\mathcal{L}_{\mathcal{D},c}$ is given as
\begin{IEEEeqnarray}{rCl}
\widehat{\mathcal{R}}_{\mathcal{D}}(\mathcal{L}_{\mathcal{D},c}) &  = & \frac{1}{N} \mathop{\mathbb{E}}_{\sigma } \left [ \sup_{l_{f_{y \mapsto z_c}} \in \mathcal{L}_{\mathcal{D},c}} \sum_{i=1}^N \sigma_i l_{f_{y \mapsto z_c}}(y^i,(z^i)_c) \right ] \\
& = & \frac{1}{N} \mathop{\mathbb{E}}_{\sigma } \left [ \sup_{f_{y \mapsto z_c} \in \mathcal{M}_{\mathcal{D},c}} \sum_{i=1}^N \sigma_i  |(z^i)_c-f_{y \mapsto z_c}(y^i)|^2   \right ].
\end{IEEEeqnarray} 
Define
\begin{IEEEeqnarray}{rCl}
u_j(f_{y \mapsto z_c}) & : = &  \sum_{i=1}^j \sigma_i  |(z^i)_c-f_{y \mapsto z_c}(y^i)|^2
\end{IEEEeqnarray} 
to express
\begin{IEEEeqnarray}{rCl}
\widehat{\mathcal{R}}_{\mathcal{D}}(\mathcal{L}_{\mathcal{D},c}) &  = & \frac{1}{N} \mathop{\mathbb{E}}_{\sigma_1,\cdots, \sigma_{N-1} } \left [ \mathop{\mathbb{E}}_{\sigma_N} \left[ \sup_{f_{y \mapsto z_c} \in \mathcal{M}_{\mathcal{D},c}} \left( u_{N-1}(f_{y \mapsto z_c}) + \sigma_N |(z^N)_c-f_{y \mapsto z_c}(y^N)|^2  \right) \right ]  \right ].   \IEEEeqnarraynumspace
\end{IEEEeqnarray}
Let $f_1,f_2 \in \mathcal{M}_{\mathcal{D},c}$ be such that 
\begin{IEEEeqnarray}{rCl}
\sup_{f_{y \mapsto z_c} \in \mathcal{M}_{\mathcal{D},c}} \left( u_{N-1}(f_{y \mapsto z_c}) +  |(z^N)_c-f_{y \mapsto z_c}(y^N)|^2  \right) & = & u_{N-1}(f_1) +  |(z^N)_c-f_1(y^N)|^2  \\
\sup_{f_{y \mapsto z_c} \in \mathcal{M}_{\mathcal{D},c}} \left( u_{N-1}(f_{y \mapsto z_c}) -  |(z^N)_c-f_{y \mapsto z_c}(y^N)|^2  \right) & = & u_{N-1}(f_2) -  |(z^N)_c-f_2(y^N)|^2. 
\end{IEEEeqnarray} 
Consider,
\begin{IEEEeqnarray}{rCl}
\nonumber \lefteqn{\mathop{\mathbb{E}}_{\sigma_N} \left[ \sup_{f_{y \mapsto z_c} \in \mathcal{M}_{\mathcal{D},c}} \left( u_{N-1}(f_{y \mapsto z_c}) + \sigma_N  |(z^N)_c-f_{y \mapsto z_c}(y^N)|^2  \right) \right]} \\
 & = & \frac{1}{2} \sup_{f_{y \mapsto z_c} \in \mathcal{M}_{\mathcal{D},c}} \left( u_{N-1}(f_{y \mapsto z_c}) +  |(z^N)_c-f_{y \mapsto z_c}(y^N)|^2  \right) \\
& & + \frac{1}{2} \sup_{f_{y \mapsto z_c} \in \mathcal{M}_{\mathcal{D},c}} \left( u_{N-1}(f_{y \mapsto z_c}) -  |(z^N)_c-f_{y \mapsto z_c}(y^N)|^2  \right). 
\end{IEEEeqnarray}
Thus,
\begin{IEEEeqnarray}{rCl}
\lefteqn{\mathop{\mathbb{E}}_{\sigma_N} \left[ \sup_{f_{y \mapsto z_c} \in \mathcal{M}_{\mathcal{D},c}} \left( u_{N-1}(f_{y \mapsto z_c}) + \sigma_N  |(z^N)_c-f_{y \mapsto z_c}(y^N)|^2  \right) \right]} \\
& = & \frac{1}{2} \left [ u_{N-1}(f_1) + u_{N-1}(f_2) \right] + \frac{1}{2}  \left [ |(z^N)_c-f_1(y^N)|^2  - |(z^N)_c-f_2(y^N)|^2 \right] \\
& = & \frac{1}{2} \left [ u_{N-1}(f_1) + u_{N-1}(f_2) \right] + \frac{1}{2} \left[ \left(f_1(y^N) + f_2(y^N) - 2(z^N)_c \right) \left( f_1(y^N) - f_2(y^N) \right) \right] .
\end{IEEEeqnarray}
Define
\begin{IEEEeqnarray}{rCl}
\eta & = & \sign(f_1(y^N) - f_2(y^N)).
\end{IEEEeqnarray}
Since $f_1,f_2 \in \mathcal{M}_{\mathcal{D},c}$, we have $|f_1(y^N)| < 1$, $|f_2(y^N)| < 1$, and also $(z^N)_c \in \{0,1\}$, leading to 
\begin{IEEEeqnarray}{rCl}
f_1(y^N) + f_2(y^N) - 2(z^N)_c & \leq & 2,
\end{IEEEeqnarray}
so that 
\begin{IEEEeqnarray}{rCl}
\nonumber \lefteqn{\mathop{\mathbb{E}}_{\sigma_N} \left[ \sup_{f_{y \mapsto z_c} \in \mathcal{M}_{\mathcal{D},c}} \left( u_{N-1}(f_{y \mapsto z_c}) + \sigma_N  |(z^N)_c-f_{y \mapsto z_c}(y^N)|^2  \right) \right]} \\
 & \leq & \frac{1}{2} \left [ u_{N-1}(f_1) + u_{N-1}(f_2) \right] + \frac{1}{2} \left[ 2 \eta  \left( f_1(y^N) - f_2(y^N) \right) \right] \\
& = & \frac{1}{2} \left [ u_{N-1}(f_1) + 2 \eta f_1(y^N)  \right] + \frac{1}{2} \left[ u_{N-1}(f_2) - 2 \eta   f_2(y^N)  \right] \\
\nonumber & \leq & \frac{1}{2} \sup_{f_{y \mapsto z_c} \in \mathcal{M}_{\mathcal{D},c}}\left( u_{N-1}(f_{y \mapsto z_c}) + 2  \eta f_{y \mapsto z_c}(y^N) \right) \\
&& + \frac{1}{2} \sup_{f_{y \mapsto z_c} \in \mathcal{M}_{\mathcal{D},c}}\left( u_{N-1}(f_{y \mapsto z_c}) - 2 \eta   f_{y \mapsto z_c}(y^N) \right) \\
& = & \mathop{\mathbb{E}}_{\sigma_N} \left[ \sup_{f_{y \mapsto z_c} \in \mathcal{M}_{\mathcal{D},c}}\left( u_{N-1}(f_{y \mapsto z_c}) + \sigma_N 2  f_{y \mapsto z_c}(y^N)  \right) \right].
\end{IEEEeqnarray}
That is,
\begin{IEEEeqnarray}{rCl}
\nonumber \lefteqn{\mathop{\mathbb{E}}_{\sigma_N} \left[ \sup_{f_{y \mapsto z_c} \in \mathcal{M}_{\mathcal{D},c}} \left( u_{N-1}(f_{y \mapsto z_c}) + \sigma_N  |(z^N)_c-f_{y \mapsto z_c}(y^N)|^2  \right) \right]} \\
 \label{eq_100420241314}& \leq & \mathop{\mathbb{E}}_{\sigma_N} \left[ \sup_{f_{y \mapsto z_c} \in \mathcal{M}_{\mathcal{D},c}}\left( u_{N-1}(f_{y \mapsto z_c}) + \sigma_N 2   f_{y \mapsto z_c}(y^N)  \right) \right].
\end{IEEEeqnarray}
In other words,
\begin{IEEEeqnarray}{rCl}
\nonumber \lefteqn{\mathop{\mathbb{E}}_{\sigma_N} \left[ \sup_{f_{y \mapsto z_c} \in \mathcal{M}_{\mathcal{D},c}} \sum_{i=1}^N \sigma_i  |(z^i)_c-f_{y \mapsto z_c}(y^i)|^2  \right]} \\
 & \leq & \mathop{\mathbb{E}}_{\sigma_N} \left[ \sup_{f_{y \mapsto z_c} \in \mathcal{M}_{\mathcal{D},c}}\left( u_{N-1}(f_{y \mapsto z_c}) + \sigma_N 2  f_{y \mapsto z_c}(y^N)  \right) \right]. 
\end{IEEEeqnarray}
That is,
\begin{IEEEeqnarray}{rCl}
\nonumber \lefteqn{\mathop{\mathbb{E}}_{\sigma_{N-1},\sigma_N} \left[ \sup_{f_{y \mapsto z_c} \in \mathcal{M}_{\mathcal{D},c}} \sum_{i=1}^N \sigma_i  |(z^i)_c-f_{y \mapsto z_c}(y^i)|^2  \right] \leq} \\
&  & \mathop{\mathbb{E}}_{\sigma_N}  \left[ \mathop{\mathbb{E}}_{\sigma_{N-1}} \left [ \sup_{f_{y \mapsto z_c} \in \mathcal{M}_{\mathcal{D},c}}\left( u_{N-2}(f_{y \mapsto z_c}) + \sigma_N 2  f_{y \mapsto z_c}(y^N)  + \sigma_{N-1}  |z^{N-1}-f_{y \mapsto z_c}(y^{N-1})|^2   \right) \right] \right] \IEEEeqnarraynumspace
\end{IEEEeqnarray}
We can follow the same procedure for $\sigma_{N-1}$, as did for $\sigma_{N}$ for deriving (\ref{eq_100420241314}), to show that
\begin{IEEEeqnarray}{rCl}
\nonumber \lefteqn{\mathop{\mathbb{E}}_{\sigma_{N-1}} \left[ \sup_{f_{y \mapsto z_c} \in \mathcal{M}_{\mathcal{D},c}} \left( u_{N-2}(f_{y \mapsto z_c}) + \sigma_N 2\bar{s}  f_{y \mapsto z_c}(y^N) + \sigma_{N-1}  |z^{N-1}-f_{y \mapsto z_c}(y^{N-1})|^2  \right) \right]}\\
& \leq & \mathop{\mathbb{E}}_{\sigma_{N-1}} \left[ \sup_{f_{y \mapsto z_c} \in \mathcal{M}_{\mathcal{D},c}}\left( u_{N-2}(f_{y \mapsto z_c}) + \sigma_N 2  f_{y \mapsto z_c}(y^N) + \sigma_{N-1} 2  f_{y \mapsto z_c}(y^{N-1})  \right) \right].
\end{IEEEeqnarray}
That is,
\begin{IEEEeqnarray}{rCl}
\nonumber \lefteqn{\mathop{\mathbb{E}}_{\sigma_{N-1},\sigma_N} \left[ \sup_{f_{y \mapsto z_c} \in \mathcal{M}_{\mathcal{D},c}} \sum_{i=1}^N \sigma_i  |(z^i)_c-f_{y \mapsto z_c}(y^i)|^2 \right]}\\
& \leq & \mathop{\mathbb{E}}_{\sigma_{N-1},\sigma_N} \left[ \sup_{f_{y \mapsto z_c} \in \mathcal{M}_{\mathcal{D},c}}\left( u_{N-2}(f_{y \mapsto z_c})  + \sigma_{N-1} 2  f_{y \mapsto z_c}(y^{N-1}) + \sigma_N 2 f_{y \mapsto z_c}(y^N)  \right) \right].
\end{IEEEeqnarray}
Proceeding in the same way, we get
\begin{IEEEeqnarray}{rCl}
\nonumber \lefteqn{\mathop{\mathbb{E}}_{\sigma_1,\cdots,\sigma_{N-1},\sigma_N} \left[ \sup_{f_{y \mapsto z_c} \in \mathcal{M}_{\mathcal{D},c}} \sum_{i=1}^N \sigma_i  |(z^i)_c-f_{y \mapsto z_c}(y^i)|^2 \right]} \\
& \leq & \mathop{\mathbb{E}}_{\sigma_1,\cdots,\sigma_{N-1},\sigma_N} \left[ \sup_{f_{y \mapsto z_c} \in \mathcal{M}_{\mathcal{D},c}}\left( 2  \sum_{i=1}^N \sigma_i f_{y \mapsto z_c}(y^{i}) \right) \right] \\
& = & 2 N \widehat{\mathcal{R}}_{\mathcal{D}}(\mathcal{M}_{\mathcal{D},c}) .
\end{IEEEeqnarray}
Hence,
\begin{IEEEeqnarray}{rCl}
\widehat{\mathcal{R}}_{\mathcal{D}}(\mathcal{L}_{\mathcal{D},c}) & \leq & 2 \widehat{\mathcal{R}}_{\mathcal{D}}(\mathcal{M}_{\mathcal{D},c}). 
\end{IEEEeqnarray} 
Now, using Theorem~\ref{result_rademacher_complexity_hypthesis_space} leads to (\ref{eq_150120241310}) and (\ref{eq_150120241311}).
\section*{Appendix E. Proof of Theorem~\ref{result_generalisation_error}}
We define a function assessing the supremum of difference of expected loss value from the empirical averaged loss value:
\begin{IEEEeqnarray}{rCl}
\phi(\mathcal{D}) & = & \sup_{f_{y \mapsto z_c}  \in \mathcal{M}_{\mathcal{D},c}} \left( \mathop{\mathbb{E}}_{(y,z) \sim \mathbb{P}_{y,z}}[l_{f_{y \mapsto z_c}}(y,z_c)] - \widehat{\mathbb{E}}_{\mathcal{D}}[l_{f_{y \mapsto z_c}}]  \right). 
\end{IEEEeqnarray}
Let $\mathcal{D}^{\prime} = \{(y^1,z^1,q^1),\cdots,(y^{i-1},z^{i-1},q^{i-1}),(\tilde{y}^i,\tilde{z}^i,\tilde{q}^i),(y^{i+1},z^{i+1},q^{i+1}),\cdots,(y^N,z^N,q^N) \}$ be a ``neighboring'' set of $\mathcal{D}$ such that $\mathcal{D}^{\prime} $ differs from $\mathcal{D}$ by only single entry, i.e., $(\tilde{y}^i,\tilde{z}^i,\tilde{q}^i) \notin \mathcal{D}$ and $(y^i,z^i,q^i) \notin \mathcal{D}^{\prime}$. As the difference of suprema can't exceed the supremum of the difference, we have
\begin{IEEEeqnarray}{rCl}
\nonumber \phi(\mathcal{D}^{\prime}) - \phi(\mathcal{D}) & \leq & \sup_{f_{y \mapsto z_c}  \in \mathcal{M}_{\mathcal{D},c}} \left( \mathop{\mathbb{E}}_{(y,z) \sim  \mathbb{P}_{y,z}}[l_{f_{y \mapsto z_c}}(y,z_c)] - \widehat{\mathbb{E}}_{\mathcal{D}^{\prime}}[l_{f_{y \mapsto z_c}}] \right. \\
&& \left. - \mathop{\mathbb{E}}_{(y,z) \sim \mathbb{P}_{y,z}}[l_{f_{y \mapsto z_c}}(y,z_c)] + \widehat{\mathbb{E}}_{\mathcal{D}}[l_{f_{y \mapsto z_c}}]  \right) \\
& = & \sup_{f_{y \mapsto z_c}  \in \mathcal{M}_{\mathcal{D},c}} \left( \widehat{\mathbb{E}}_{\mathcal{D}}[l_{f_{y \mapsto z_c}}]  - \widehat{\mathbb{E}}_{\mathcal{D}^{\prime}}[l_{f_{y \mapsto z_c}}]  \right) \\
& = & \sup_{f_{y \mapsto z_c}  \in \mathcal{M}_{\mathcal{D},c}} \frac{l_{f_{y \mapsto z_c}}(y^i,(z^i)_c)-l_{f_{y \mapsto z_c}}(\tilde{y}^i,(\tilde{z}^i)_c)}{N} \\
& = & \sup_{f_{y \mapsto z_c} \in \mathcal{M}_{\mathcal{D},c}}\frac{|(z^i)_c-f_{y \mapsto z_c}(y^i)|^2 - |(\tilde{z}^i)_c-f_{y \mapsto z_c}(\tilde{y}^i)|^2 }{N} \\
\label{eq_110420241814}& \leq & \frac{1}{N},
\end{IEEEeqnarray}
where (\ref{eq_110420241814}) follows from the facts that $(z^i)_c \in \{0,1\}$ and $f_{y \mapsto z_c}(y^i) \in [0,1]$. Similarly, we can obtain
\begin{IEEEeqnarray}{rCl}
\phi(\mathcal{D}) - \phi(\mathcal{D}^{\prime})  & \leq &  \frac{1}{N}.
\end{IEEEeqnarray}
Thus
\begin{IEEEeqnarray}{rCl}
|\phi(\mathcal{D}) - \phi(\mathcal{D}^{\prime})|  & \leq &  \frac{1}{N}.
\end{IEEEeqnarray} 
Thus, $\phi$ satisfies the bounded differences property with bound $1/N$, and therefore by McDiarmid's inequality, for any $\epsilon > 0$, with probability at most $\exp(-2N\epsilon^2)$, the following holds:   
\begin{IEEEeqnarray}{rCl}
\phi(\mathcal{D}) - \mathop{\mathbb{E}}_{\mathcal{D} \sim (\mathbb{P}_{y,z,q})^N}[\phi(\mathcal{D})] & \geq & \epsilon .
\end{IEEEeqnarray} 
That is, with probability at most $\delta > 0$, the following holds:
\begin{IEEEeqnarray}{rCl}
\phi(\mathcal{D}) - \mathop{\mathbb{E}}_{\mathcal{D} \sim (\mathbb{P}_{y,z,q})^N}[\phi(\mathcal{D})] & \geq & \sqrt{\frac{\log(1/\delta)}{2N}}. 
\end{IEEEeqnarray}
In other words, with probability at least $1-\delta$, the following holds:
\begin{IEEEeqnarray}{rCl}
\phi(\mathcal{D})  & \leq & \mathop{\mathbb{E}}_{\mathcal{D} \sim (\mathbb{P}_{y,z,q})^N}[\phi(\mathcal{D})] + \sqrt{\frac{\log(1/\delta)}{2N}}. 
\end{IEEEeqnarray}
Let $\tilde{\mathcal{D}} = \left \{(\tilde{y}^i,\tilde{z}^i,\tilde{q}^i) \sim \mathbb{P}_{y,z,q}\; \mid \;i \in \{1,2,\cdots,N\} \right \} $ be another set of i.i.d. samples. Consider
\begin{IEEEeqnarray}{rCl}
\mathop{\mathbb{E}}_{\mathcal{D} \sim (\mathbb{P}_{y,z,q})^N}[\phi(\mathcal{D})] & = &  \mathop{\mathbb{E}}_{\mathcal{D} \sim (\mathbb{P}_{y,z,q})^N}\left[ \sup_{f_{y \mapsto z_c} \in \mathcal{M}_{\mathcal{D},c}} \left( \mathop{\mathbb{E}}_{(y,z) \sim \mathbb{P}_{y,z}}[l_{f_{y \mapsto z_c}}(y,z_c)] - \widehat{\mathbb{E}}_{\mathcal{D}}[l_{f_{y \mapsto z_c}}]  \right) \right] \IEEEeqnarraynumspace \\
\label{eq_110420241944}& = & \mathop{\mathbb{E}}_{\mathcal{D} \sim (\mathbb{P}_{y,z,q})^N}\left[ \sup_{f_{y \mapsto z_c} \in \mathcal{M}_{\mathcal{D},c}} \left( \mathop{\mathbb{E}}_{\tilde{\mathcal{D}} \sim (\mathbb{P}_{y,z,q})^N}\left [ \widehat{\mathbb{E}}_{\tilde{\mathcal{D}}}[l_{f_{y \mapsto z_c}}] - \widehat{\mathbb{E}}_{\mathcal{D}}[l_{f_{y \mapsto z_c}}] \right ]  \right) \right],
\end{IEEEeqnarray}
where we have used the fact that 
\begin{IEEEeqnarray}{rCCCl}
\mathop{\mathbb{E}}_{\tilde{\mathcal{D}} \sim (\mathbb{P}_{y,z,q})^N}\left [ \widehat{\mathbb{E}}_{\tilde{\mathcal{D}}}[l_{f_{y \mapsto z_c}}] \right ] & = & \mathop{\mathbb{E}}_{(y,z,q) \sim \mathbb{P}_{y,z,q}}[l_{f_{y \mapsto z_c}}(y,z_c)] & = & \mathop{\mathbb{E}}_{(y,z) \sim \mathbb{P}_{y,z}}[l_{f_{y \mapsto z_c}}(y,z_c)].
\end{IEEEeqnarray}
It follows from (\ref{eq_110420241944}) that
\begin{IEEEeqnarray}{rCl}
\nonumber \lefteqn{\mathop{\mathbb{E}}_{\mathcal{D} \sim (\mathbb{P}_{y,z,q})^N}[\phi(\mathcal{D})]} \\
 & \leq  & \mathop{\mathbb{E}}_{\mathcal{D} \sim (\mathbb{P}_{y,z,q})^N,\tilde{\mathcal{D}} \sim (\mathbb{P}_{y,z,q})^N}\left[ \sup_{f_{y \mapsto z_c} \in \mathcal{M}_{\mathcal{D},c}} \left(  \widehat{\mathbb{E}}_{\tilde{\mathcal{D}}}[l_{f_{y \mapsto z_c}}] - \widehat{\mathbb{E}}_{\mathcal{D}}[l_{f_{y \mapsto z_c}}]   \right) \right]  \IEEEeqnarraynumspace \\
& = & \mathop{\mathbb{E}}_{\mathcal{D} \sim (\mathbb{P}_{y,z,q})^N,\tilde{\mathcal{D}} \sim (\mathbb{P}_{y,z,q})^N}\left[ \sup_{f_{y \mapsto z_c} \in \mathcal{M}_{\mathcal{D},c}} \left( \frac{1}{N} \sum_{i=1}^N \left(l_{f_{y \mapsto z_c}}\left(\tilde{y}^i,(\tilde{z}^i)_c\right) - l_{f_{y \mapsto z_c}}\left(y^i,(z^i)_c\right) \right)  \right) \right] \IEEEeqnarraynumspace \\
\label{eq_202401011743}& = & \mathop{\mathbb{E}}_{\mathcal{D} \sim (\mathbb{P}_{y,z,q})^N,\tilde{\mathcal{D}} \sim (\mathbb{P}_{y,z,q})^N,\sigma }\left[ \sup_{f_{y \mapsto z_c} \in \mathcal{M}_{\mathcal{D},c}} \left( \frac{1}{N} \sum_{i=1}^N \sigma_i \left( l_{f_{y \mapsto z_c}}\left(\tilde{y}^i,(\tilde{z}^i)_c\right) - l_{f_{y \mapsto z_c}}\left(y^i,(z^i)_c\right) \right)  \right) \right], \IEEEeqnarraynumspace
\end{IEEEeqnarray}
where we have considered the facts that
\begin{itemize}
\item $\sigma_1,\cdots,\sigma_N$ are Rademacher variables (i.e. taking values in $\{ -1,1\}$ with probability equal to 1/2), and 
\item we have
\begin{IEEEeqnarray}{rCl}
\nonumber \lefteqn{\mathop{\mathbb{E}}_{\mathcal{D} \sim (\mathbb{P}_{y,z,q})^N,\tilde{\mathcal{D}} \sim (\mathbb{P}_{y,z,q})^N }\left[ \sup_{f_{y \mapsto z_c} \in \mathcal{M}_{\mathcal{D},c}} \left( \frac{1}{N} \sum_{i=1}^N \left(l_{f_{y \mapsto z_c}}\left(\tilde{y}^i,(\tilde{z}^i)_c\right) - l_{f_{y \mapsto z_c}}\left(y^i,(z^i)_c\right) \right)  \right) \right]} \\
& = & \mathop{\mathbb{E}}_{\mathcal{D} \sim (\mathbb{P}_{y,z,q})^N,\tilde{\mathcal{D}} \sim (\mathbb{P}_{y,z,q})^N}\left[ \sup_{f_{y \mapsto z_c} \in \mathcal{M}_{\mathcal{D},c}} \left( \frac{1}{N} \sum_{i=1}^N \left(  l_{f_{y \mapsto z_c}}(y^i,(z^i)_c)  - l_{f_{y \mapsto z_c}}(\tilde{y}^i,(\tilde{z}^i)_c) \right) \right) \right]. \IEEEeqnarraynumspace
\end{IEEEeqnarray}
\end{itemize}
It follows from (\ref{eq_202401011743}) that
\begin{IEEEeqnarray}{rCl}
\nonumber \mathop{\mathbb{E}}_{\mathcal{D} \sim (\mathbb{P}_{y,z,q})^N}[\phi(\mathcal{D})] & \leq  & \mathop{\mathbb{E}}_{\tilde{\mathcal{D}} \sim (\mathbb{P}_{y,z,q})^N,\sigma }\left[ \sup_{f_{y \mapsto z_c} \in \mathcal{M}_{\mathcal{D},c}} \left( \frac{1}{N} \sum_{i=1}^N \sigma_i  l_{f_{y \mapsto z_c}}\left(\tilde{y}^i,(\tilde{z}^i)_c\right)   \right) \right] \\
&& + \mathop{\mathbb{E}}_{\mathcal{D} \sim (\mathbb{P}_{y,z,q})^N,\sigma }\left[ \sup_{f_{y \mapsto z_c} \in \mathcal{M}_{\mathcal{D},c}} \left( \frac{1}{N} \sum_{i=1}^N - \sigma_i  l_{f_{y \mapsto z_c}}\left(y^i,(z^i)_c\right)   \right) \right].
\end{IEEEeqnarray}
Since $\sigma_i$ and $-\sigma_i$ are distributed identically, we have
\begin{IEEEeqnarray}{rCl}
\mathop{\mathbb{E}}_{\mathcal{D} \sim (\mathbb{P}_{y,z,q})^N}[\phi(\mathcal{D})] & \leq  &  2 \mathop{\mathbb{E}}_{\mathcal{D} \sim (\mathbb{P}_{y,z,q})^N,\sigma }\left[ \sup_{f_{y \mapsto z_c} \in \mathcal{M}_{\mathcal{D},c}} \left( \frac{1}{N} \sum_{i=1}^N  \sigma_i  l_{f_{y \mapsto z_c}}\left(y^i,(z^i)_c\right)   \right) \right] \\
& = & 2 \mathop{\mathbb{E}}_{\mathcal{D} \sim (\mathbb{P}_{y,z,q})^N,\sigma }\left[ \sup_{l_{f_{y \mapsto z_c}} \in \mathcal{L}_{\mathcal{D},c}} \left( \frac{1}{N} \sum_{i=1}^N  \sigma_i  l_{f_{y \mapsto z_c}}\left(y^i,(z^i)_c\right)   \right) \right] \\
& = & 2 \mathop{\mathbb{E}}_{\mathcal{D} \sim (\mathbb{P}_{y,z,q})^N }\left[ \widehat{\mathcal{R}}_{\mathcal{D}}(\mathcal{L}_{\mathcal{D},c}) \right] \\
& \leq & \frac{4 }{\sqrt{N}},
\end{IEEEeqnarray}
where we have used Lemma~\ref{result_rademacher_complexity_loss_space}. Thus, with probability at least $1-\delta$, we have 
\begin{IEEEeqnarray}{rCl}
\phi(\mathcal{D})  & \leq &  \frac{4   }{\sqrt{N}} + \sqrt{\frac{\log(1/\delta)}{2N}}. 
\end{IEEEeqnarray}
Therefore, for any $f_{y \mapsto z_c} \in \mathcal{M}_{\mathcal{D},c}$, we have with probability at least $1-\delta$,
\begin{IEEEeqnarray}{rCl}
\label{eq_170220241926}\mathop{\mathbb{E}}_{(y,z) \sim \mathbb{P}_{y,z}}[l_{f_{y \mapsto z_c}}(y,z_c)] & \leq & \frac{1}{N}\sum_{i=1}^N l_{f_{y \mapsto z_c}}(y^i,(z^i)_c) +  \frac{4  }{\sqrt{N}} + \sqrt{\frac{\log(1/\delta)}{2N}}.
\end{IEEEeqnarray} 
\section*{Appendix F. Proof of Theorem~\ref{theorem_learning_solution}}
Problem~\ref{problem_learning} can be formulated as
 \begin{IEEEeqnarray}{rCl}
 f_{y \mapsto z_c}^* & = & \sum_{q=1}^Q \sum_{i=\mathrm{I}_1^{c,q}}^{\mathrm{I}_{|\mathcal{I}^{c,q}|}^{c,q}} \alpha_{c,i}^* \mathcal{K}_{c}(\cdot,y^i),\; \mbox{where}
  \end{IEEEeqnarray} 
 \begin{IEEEeqnarray}{rCl}
\nonumber \lefteqn{\left\{\alpha_{c,i}^*\; \mid \;i \in \mathop{\cup}_{q=1}^Q \mathcal{I}^{c,q} \right\}} \\
\label{eq_120420241312} & = & \mathop{\argmin}_{\displaystyle \left\{\alpha_{c,i}\; \mid \; \alpha_{c,i} \in [0,1],\; \sum_{q=1}^Q \sum_{i=\mathrm{I}_1^{c,q}}^{\mathrm{I}_{|\mathcal{I}^{c,q}|}^{c,q}}  \alpha_{c,i} = 1 \right \}} \; \mathop{\mathbb{E}}_{(y,z) \sim \mathbb{P}_{y,z}}\left [ \left |z_c-\sum_{q=1}^Q \sum_{i=\mathrm{I}_1^{c,q}}^{\mathrm{I}_{|\mathcal{I}^{c,q}|}^{c,q}} \alpha_{c,i} \mathcal{K}_{c}(y,y^i)\right |^2 \right ] \IEEEeqnarraynumspace
 \end{IEEEeqnarray} 
As a result of Assumption~\ref{assumption_1}, 
 \begin{IEEEeqnarray}{rCl}
\label{eq_160120241028}f_{y \mapsto z_c}^*(y) & \approx & \exp\left(-\frac{1}{p} \Gamma_{\mathcal{G}_c}(y)\right) \sum_{q=1}^Q \sum_{i=\mathrm{I}_1^{c,q}}^{\mathrm{I}_{|\mathcal{I}^{c,q}|}^{c,q}} \alpha_{c,i}^*.
   \end{IEEEeqnarray} 
Since
 \begin{IEEEeqnarray}{rCl}
\sum_{q=1}^Q \sum_{i=\mathrm{I}_1^{c,q}}^{\mathrm{I}_{|\mathcal{I}^{c,q}|}^{c,q}} \alpha_{c,i}^*&=& 1,
   \end{IEEEeqnarray} 
we get (\ref{eq_150120241814}).  
\section*{Appendix G. Proof of Theorem~\ref{result_generalisation_error_2}}
Since $f_{y \mapsto z_c}^* \in \mathcal{M}_{\mathcal{D},c}$, it follows from Theorem~\ref{result_generalisation_error} that we have with probability at least $1-\delta$ for any $\delta \in (0,1)$,
\begin{IEEEeqnarray}{rCl}
\mathop{\mathbb{E}}_{(y,z) \sim \mathbb{P}_{y,z}}\left [\left |z_c-f_{y \mapsto z_c}^*(y) \right|^2 \right ]
& \leq & \frac{1}{N}\sum_{i=1}^N \left |(z^i)_c-f_{y \mapsto z_c}^*(y^i) \right|^2  +  \frac{4  }{\sqrt{N}} + \sqrt{\frac{\log(1/\delta)}{2N}}.
\end{IEEEeqnarray} 
Due to Assumption~\ref{assumption_1} and Assumption~\ref{assumption_2}, 
  \begin{IEEEeqnarray}{rCl}
 f_{y \mapsto z_c}^*(y^i)&\approx& (z^i)_c,\;\forall i \in \{1,2,\cdots,N \}.
   \end{IEEEeqnarray} 
That is,
  \begin{IEEEeqnarray}{rCl}
\frac{1}{N} \sum_{i=1}^N \left |(z^i)_c-f_{y \mapsto z_c}^*(y^i) \right|^2 & \approx & 0.
   \end{IEEEeqnarray}    
Hence, the result follows.
\section*{Appendix H. Proof of Theorem~\ref{result_generalisation_error_3}}
Consider
  \begin{IEEEeqnarray}{rCl}
\nonumber \lefteqn{\mathop{\mathbb{E}}_{(y,z) \sim \mathbb{P}_{y,z}}\left [\left |z_c-f_{y \mapsto z_c}^*(y) \right|^2 \right ] - \mathop{\mathbb{E}}_{(y,z) \sim \mathbb{P}_{y,z}}\left [\left |z_c - \mathbb{P}_{z | y}(z_c = 1 | y)  \right|^2 \right ]} \\
& = & \mathop{\mathbb{E}}_{(y,z) \sim \mathbb{P}_{y,z}}\left [ |f_{y \mapsto z_c}^*(y)|^2 - |\mathbb{P}_{z | y}(z_c = 1 | y)|^2 - 2 z_c \left(f_{y \mapsto z_c}^*(y) - \mathbb{P}_{z | y}(z_c = 1 | y) \right) \right ] \IEEEeqnarraynumspace \\
\nonumber & = & \mathop{\mathbb{E}}_{y \sim \mathbb{P}_{y}}\left [ |f_{y \mapsto z_c}^*(y)|^2\right] - \mathop{\mathbb{E}}_{y \sim \mathbb{P}_{y}}\left [ |\mathbb{P}_{z | y}(z_c = 1 | y)|^2\right] \\
& & - 2 \mathop{\mathbb{E}}_{y \sim \mathbb{P}_{y}}\left[ \mathbb{P}_{z | y}(z_c = 1 | y) \left(f_{y \mapsto z_c}^*(y) - \mathbb{P}_{z | y}(z_c = 1 | y) \right)  \right] \\
& = & \mathop{\mathbb{E}}_{y \sim \mathbb{P}_{y}}\left [ |f_{y \mapsto z_c}^*(y) -  \mathbb{P}_{z | y}(z_c = 1 | y) |^2\right].
   \end{IEEEeqnarray} 
Thus,
\begin{IEEEeqnarray}{rCl}
\label{eq_130420241527} \mathop{\mathbb{E}}_{y \sim \mathbb{P}_{y}}\left [ |f_{y \mapsto z_c}^*(y) -  \mathbb{P}_{z | y}(z_c = 1 | y) |^2\right] & \leq & \mathop{\mathbb{E}}_{(y,z) \sim \mathbb{P}_{y,z}}\left [\left |z_c-f_{y \mapsto z_c}^*(y) \right|^2 \right ].
\end{IEEEeqnarray} 
Hence, the results follow by using (\ref{eq_130420241527}) in Theorem~\ref{result_generalisation_error} and Theorem~\ref{result_generalisation_error_2}. 
\section*{Appendix I. Proof of Theorem~\ref{theorem_deterministic_analysis}}
It follows from (\ref{eq_150120241844}) that
\begin{IEEEeqnarray}{rCl}
\Gamma_{\mathcal{G}_c}(y) & = & \Gamma_{\mathcal{A}_{\mathbf{Y}^{c,q^*(y)}}}(y). 
\end{IEEEeqnarray}
Using (\ref{eq_100120231432}), we get 
\begin{IEEEeqnarray}{rCl}
\label{eq_030520242105}\Gamma_{\mathcal{G}_c}(y) & < &  \left( 1 + \frac{p \left|\mathcal{I}^{c,q^*(y)}\right|^2}{2 \left\|\mathbf{Y}^{c,q^*(y)} \right\|_F^2} \right) \left \| \left[\begin{IEEEeqnarraybox*}[][c]{,c/c/c,} y - y^{\displaystyle \mathrm{I}_1^{c,q^*(y)}} & \cdots & y - y^{\displaystyle \mathrm{I}_{\left|\mathcal{I}^{c,q^*(y)}\right|}^{c,q^*(y)}} \end{IEEEeqnarraybox*} \right]   \right \|_2.
\end{IEEEeqnarray}
Using (\ref{eq_030520242105}) in (\ref{eq_150120241814}), we get (\ref{eq_030520242103}). 
\vskip 0.2in
\bibliography{sample}
\bibliographystyle{theapa}

\end{document}